\documentclass[10pt,twocolumn,letterpaper]{article}

\usepackage{wacv}              

\usepackage{graphicx}
\usepackage{amsmath}
\usepackage{amssymb}
\usepackage{booktabs}

\usepackage[pagebackref,breaklinks,colorlinks]{hyperref}
\usepackage{orcidlink}
\usepackage{colortbl}
\usepackage{float}
\usepackage{bbding}
\usepackage{pifont}
\usepackage{multirow}
\usepackage{amsmath}
\usepackage{bm}
\usepackage{amsfonts}
\usepackage{enumitem}
\usepackage{caption}
\usepackage{wrapfig}

\usepackage[capitalize]{cleveref}
\crefname{section}{Sec.}{Secs.}
\Crefname{section}{Section}{Sections}
\Crefname{table}{Table}{Tables}
\crefname{table}{Tab.}{Tabs.}

\def\wacvPaperID{1073} 
\def\confName{WACV}
\def\confYear{2025}

\definecolor{citecolor}{RGB}{0, 113, 188}
\definecolor{myforestgreen}{RGB}{34, 200, 34}
\definecolor{firebrick}{rgb}{0.7, 0.13, 0.13}
\definecolor{darkpastelgreen}{rgb}{0.01, 0.75, 0.24}
\definecolor{deepskyblue}{rgb}{0.0, 0.75, 1.0}
\definecolor{mypink2}{rgb}{.99,.96,.98}
\definecolor{mypink1}{rgb}{.99,.93,.98}
\definecolor{mypink}{rgb}{.99,.90,.98}
\definecolor{mygray}{rgb}{.95,.95,.95}
\definecolor{lv14}{rgb}{0.5,0.5,0.5}
\definecolor{tabvline}{HTML}{a8a495}
\definecolor{prompt_blue}{HTML}{1f78b4}
\definecolor{prompt_red}{HTML}{d45c43}
\definecolor{green_im}{rgb}{0.0, 0.5, 0.0}

\newcommand{\stdvuno}[1]{\footnotesize{\color{black}(#1)} {\color{black}}}
\newcommand{\stdvun}[1]{\footnotesize\textcolor{black}{(#1)} \small\textcolor{myforestgreen}{$\uparrow$}}
\newcommand{\stdvuru}[1]{\footnotesize\textcolor{black}{({#1})} \small\textcolor{red}{$\downarrow$}}
\newcommand{\stdvueq}[1]{\footnotesize\textcolor{black}{({#1})} \textcolor{black}{$\approx$}}

\begin{document}

\title{Test-Time Low Rank Adaptation via Confidence Maximization\\ for Zero-Shot Generalization of Vision-Language Models}

\author{Raza Imam\quad Hanan Gani\quad Muhammad Huzaifa\quad Karthik Nandakumar\\
Mohamed bin Zayed University of Artificial Intelligence (MBZUAI), UAE\\
{\tt\small \{raza.imam,hanan.ghani,muhammad.huzaifa,karthik.nandakumar\}@mbzuai.ac.ae}
}
\maketitle

\begin{abstract}
  The conventional \texttt{modus operandi} for adapting pre-trained vision-language models (VLMs) during test-time involves tuning learnable prompts, \ie, test-time prompt tuning.
  This paper introduces \textbf{T}est-\textbf{T}ime \textbf{L}ow-rank adaptation (\textbf{TTL}) as an alternative to prompt tuning for zero-shot generalization of large-scale VLMs. Taking inspiration from recent advancements in efficiently fine-tuning large language models, TTL offers a test-time parameter-efficient adaptation approach that updates the attention weights of the transformer encoder by maximizing prediction confidence. The self-supervised confidence maximization objective is specified using a \textbf{weighted entropy loss} that enforces consistency among predictions of augmented samples. TTL introduces only a small amount of trainable parameters for low-rank adapters in the model space while keeping the prompts and backbone frozen. Extensive experiments on a variety of natural distribution and cross-domain tasks show that TTL can outperform other techniques for test-time optimization of VLMs in \texttt{strict zero-shot} settings. Specifically, TTL outperforms test-time prompt tuning baselines with a significant improvement on average. Our code is available at at \href{https://github.com/Razaimam45/TTL-Test-Time-Low-Rank-Adaptation}{https://github.com/Razaimam45/TTL-Test-Time-Low-Rank-Adaptation}.

\end{abstract}
\vspace{-0.25cm}
\section{Introduction}
\label{sec:intro}
In recent years, foundational vision-language models (VLMs) such as CLIP \cite{chen2022contrastive} have significantly transformed the landscape of computer vision by demonstrating remarkable proficiency in encoding diverse tasks and concepts. Trained on extensive datasets comprising millions of image-text pairs, these models exhibit decent generalizability across a spectrum of tasks. However, the process of adapting these models for specific downstream tasks through \textit{fine-tuning} often results in a compromise on their inherent generalization capabilities \cite{kumar2022fine,wortsman2022robust}. 
To address this challenge, recent works propose the incorporation of learnable prompts into the CLIP model, either in the textual \cite{zhou2022coop,zhou2022cocoop,khattak2024learning} or visual \cite{jia2022visual} branch, or both \cite{khattakMaPLe,khattak2023self}. This allows for fine-tuning only the added prompts using a few samples from the target distribution, while keeping the rest of the model frozen. While this approach has been quite effective, fine-tuning on domain-specific data inevitably diminishes the VLM's ability to generalize to unseen domains. 

\begin{figure}[!t]
    \centering
    \begin{subfigure}[b]{0.24\textwidth}
        \includegraphics[width=\textwidth]{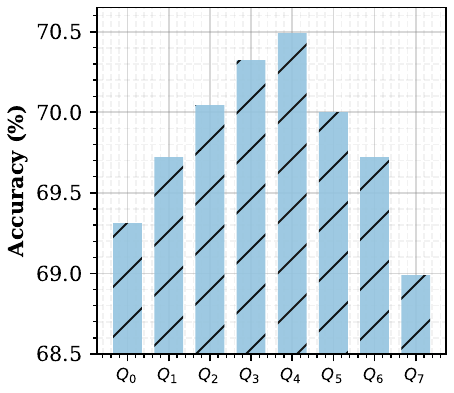}
        \caption{Entropy level \textit{vs.} Accuracy}
        \label{fig:entropy_octiles}
    \end{subfigure}
    \begin{subfigure}[b]{0.22\textwidth}
        \includegraphics[width=\textwidth]{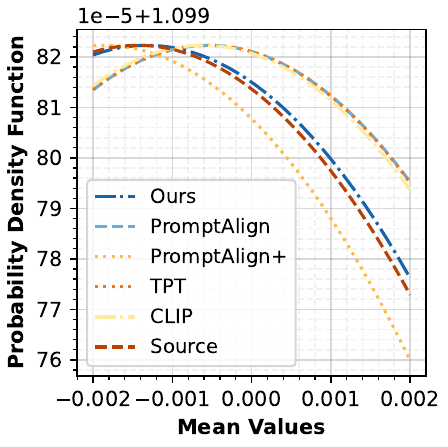}
        \caption{Visual feature statistics}
        \label{fig:gaussian}
    \end{subfigure}
        \caption{(a) Entropy corresponding to 8 different octiles result in different performance for Flowers102. (b) TTL implicitly align features such that the mean embeddings of test samples better align with that of source data (LAION) on which CLIP \cite{radford2021learning} is trained.}
\vspace{-0.4cm}
\end{figure}

Test-Time Prompt Tuning (TPT) \cite{shu2022test} was introduced as an alternative to few-shot prompt learning, where the prompts are updated dynamically on the fly for each test sample. However, TPT overlooks the \textit{distribution shift} between the training data of the CLIP model and the test samples, resulting in a subpar performance.
To address the distribution shift, PromptAlign \cite{hassan2024align} attempts to align the first-order statistics of test sample with the training data of the CLIP model. However, this approach necessitates access to a proxy dataset mimicking the distribution of CLIP training data. Additionally, it necessitates the use of pre-trained prompts for initialization, which compromises the \textit{strict zero-shot} assumption. Moreover, the token alignment achieved by PromptAlign is not as precise as that of our method, as illustrated in Figure \ref{fig:gaussian}.


\begin{figure*}[t]
    \centering
    \begin{subfigure}[b]{0.215\textwidth}
        \includegraphics[width=\textwidth]{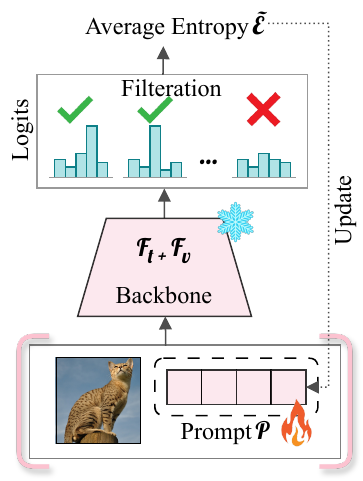}
        \caption{Existing ZS methods}
        \label{fig:existing_methods}
    \end{subfigure}
    \begin{subfigure}[b]{0.225\textwidth}
        \includegraphics[width=\textwidth]{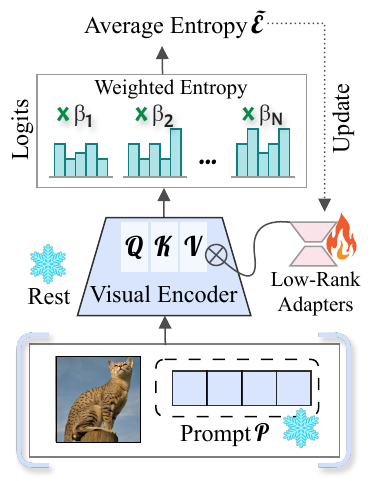}
        \caption{TTL (Ours)}
        \label{fig:ours_overview}
    \end{subfigure}
    \begin{subfigure}[b]{0.275\textwidth}
        \includegraphics[width=\textwidth]{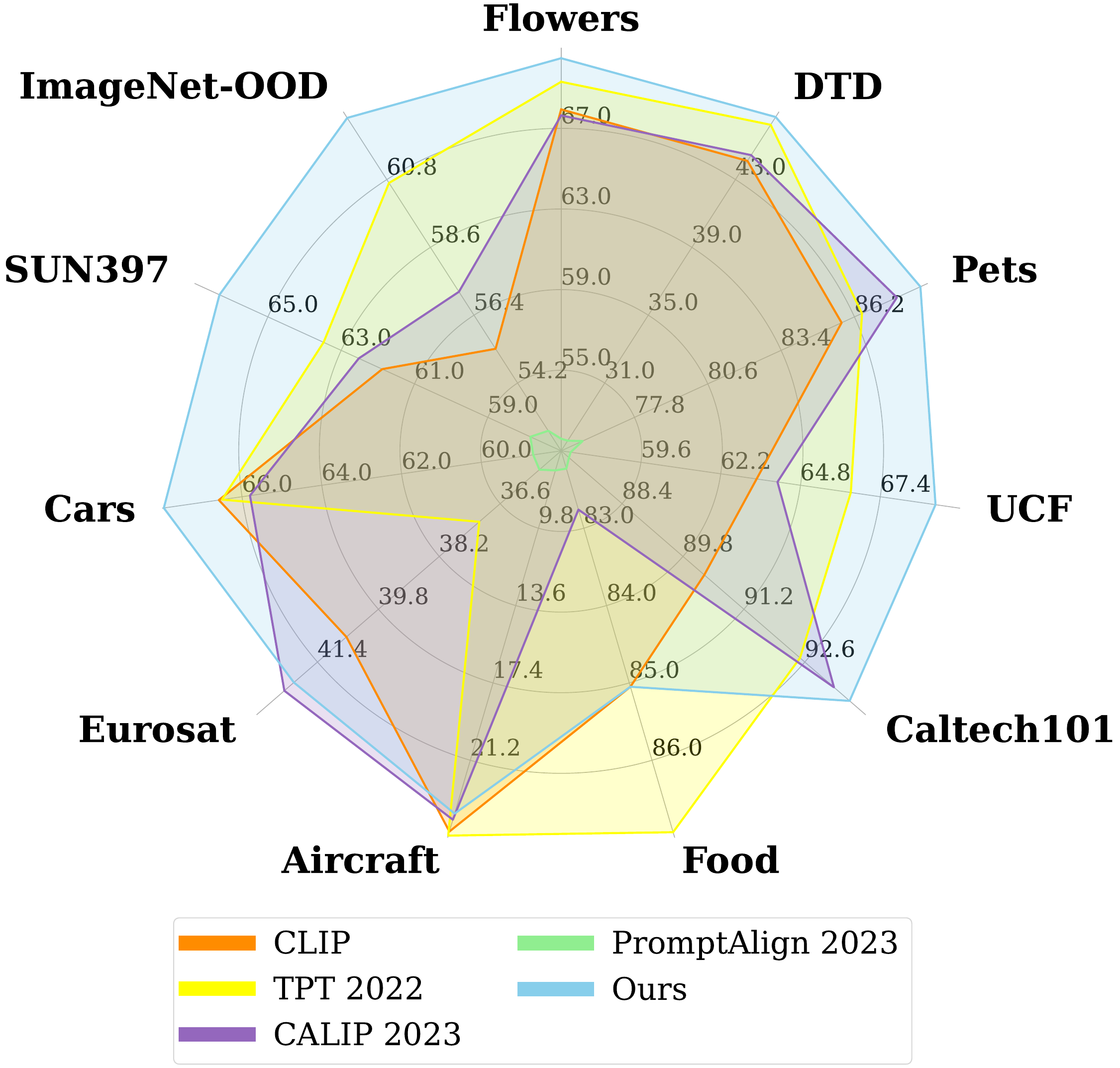}
        \caption{Results on visual classification tasks}
        \label{fig:star_plot}
    \end{subfigure}
    \hspace{0.2cm}
    \begin{minipage}[b]{0.25\textwidth}
        \caption{\textbf{TTL \textit{vs.} other zero-shot optimization methods.} (a) Current methods \cite{shu2022test, feng2023diverse, hassan2024align} update prompts during inference using self-entropy. (b) TTL introduces low-rank learnable weight matrices at the attention layer of the vision encoder to update the model weights using weighted entropy. (c) TTL outperforms existing baselines across Out-of-Distribution and Cross-Dataset while using less than 0.1\% of all model parameters.}
    \label{fig:overview}    
    \end{minipage}
\vspace{-0.4cm}  
\end{figure*}

To address the above limitations of existing test-time adaptation methods, we introduce \textbf{T}est-\textbf{T}ime \textbf{L}ow-rank adaptation (\textbf{TTL}), a parameter-efficient test-time adaptation strategy for VLMs like CLIP. TTL \textit{eliminates} the need for source data distribution during adaptation or pre-trained weights for initialization (Figure \ref{fig:overview}). Originally designed for adapting Large Language Models (LLMs) to new domains, low rank adaptation (LoRA) \cite{hu2021lora} has been extensively applied in various multi-modal and generative computer vision tasks \cite{NEURIPS2023_c1f7b1ed,lai2023lisa,chen2023minigpt,liu2023aligning,Cascante-Bonilla_2023_ICCV,Xu_2023_ICCV,NEURIPS2023_fc65fab8,guo2023animatediff,blattmann2023stable,NEURIPS2023_faacb7a4}. LoRA has two main advantages compared to prompt tuning \cite{chen2022revisiting}. Firstly, LoRA is generally more effective in low-resource (limited data availability) settings. During test-time adaptation, we have only a single unlabeled test sample available to update the model. Moreover, to minimize the overall time required for inference, only a very limited number of model updates (typically only one) are possible during test-time. Again, LoRA is known to be more stable than prompts in this scenario. It must be highlighted that our work marks \textit{the first exploration of LoRA for test-time adaptation based on a single test sample for zero-shot generalization}.


Additionally, we introduce a confidence maximization objective that replaces the conventional entropy loss used in \cite{shu2022test, hassan2024align} with a new weighted entropy loss. Existing studies \cite{beery2018recognition,geirhos2020shortcut,zhou2021examining} highlight the tendency of deep neural networks to leverage both spurious and semantically meaningful features, leading to diminished performance when spurious correlations are prevalent. Hence, relying solely on entropy for confidence estimation may not be consistently reliable under distribution shifts, as it cannot distinguish whether the model is focusing on spurious features. As shown in Figure \ref{fig:entropy_octiles}, a low entropy value is not a guarantee for correct prediction. Hence, in this work, \textit{we propose a weighted entropy loss that assigns relative weights to the different augmentations, while encouraging consistent high-confidence predictions for these augmentations}. Through empirical validation, we demonstrate the sub-optimality of using standard entropy loss to update parameters at test-time and showcase the advantages of optimizing our weighted entropy loss.
In summary, our contributions are as follows:
\vspace{-0.20cm}
\begin{itemize}[left=0pt]
    \item We introduce \textbf{T}est-\textbf{T}ime \textbf{L}oRA (\textbf{TTL}), a parameter-efficient scheme for low-rank adaptation of VLMs at test-time without relying on source data statistics or pre-trained prompts.
    \vspace{-0.20cm}
    \item We propose a weighted entropy loss that introduces a confidence maximization objective for updating parameters at test-time, showcasing its superior performance compared to the conventional entropy loss.
    \vspace{-0.20cm}
    \item We conduct extensive experiments and show that TTL achieves 7.49\% improvement on average over the baseline CLIP and 2.11\% over the best baseline for domain generalization. For cross-dataset transfer, TTL exhibits 1.40\% improvement over the baseline.
\end{itemize}



    

\section{Related Work}

\noindent\textbf{Test-Time Adaptation (TTA)}:
TTA \cite{sun2020test, wang2020tent, ttt_new} aims to bridge the distribution gap between the train and test data distributions at test time. 
While TPT \cite{shu2022test} and CALIP \cite{guo2023calip} first explored zero-shot enhancement of pre-trained VLMs, TPT relies on test-time prompt tuning, struggling with explicit alignment of pre-training and test data distributions. CALIP utilizes a parameter-free attention module for cross-modal features. PromptAlign \cite{hassan2024align} builds on TPT and aligns distribution statistics by pre-training the learnable prompts using training data, deviating from the \textit{strict zero-shot} assumption. DiffTPT \cite{feng2023diverse} employs an external diffusion model for diverse data augmentation but is impractical due to complexity of dependence on external diffusion model. In contrast, our approach efficiently updates model parameters in a single step, focusing on adapting the visual encoder of CLIP with out-of-distribution samples at test time, without relying on pre-trained weights or external sources.
\vspace{0.10cm}

\noindent\textbf{Fine-tuning for Large Vision-Language Models}: 
Having been pre-trained in a self-supervised manner on vast image-text pairs, VLMs like CLIP \cite{radford2021learning} and ALIGN \cite{jia2021scaling} have demonstrated good generalizability. However, efficiently adapting them to downstream tasks with limited data remains challenging. CoOp \cite{zhou2022coop} proposes to fine-tune CLIP by learning a set of prompts in the text encoder. CoCoOp \cite{zhou2022cocoop} highlights the inferior generalization capability of CoOp and conditions the text prompt tokens on image embeddings on the fly. MaPLe \cite{khattakMaPLe} jointly learns deep prompts at both vision and text encoders. Despite these advancements, existing methods often rely on pre-trained weights, posing challenges in real-world scenarios where no such training data from the target domain is available. In contrast, our work utilizes LoRA \cite{hu2021lora}, initialized from scratch, to adapt attention layers of the visual encoder at test time for addressing distribution shifts in out-of-domain recognition tasks.
\vspace{0.10cm}

\noindent\textbf{Entropy Minimization}:
The primary challenge of TTA is limited access to samples from the test dataset during online updates, which causes error accumulation. To mitigate this issue, TTA methods have utilized the entropy of model predictions as a confidence metric. TPT \cite{shu2022test} attempts to select the augmented samples that have minimum entropy. The need to have a batch of samples by generating multiple views via augmentations at test time is eliminated in \cite{zhang2022memo}. Motivated by TENT \cite{wang2020tent} and EATA \cite{niu2022efficient}, recently \cite{lee2023entropy} shows that entropy alone as a measure of confidence is insufficient for TTA, and propose DeYO which leverages a confidence metric called PLPD and entropy together. While effective for natural datasets, the cross-dataset performance is still a unresolved problem, which we attempt to solve using the weighted entropy loss.


\section{Methodology}
\subsection{Preliminaries}
\noindent\textbf{Contrastive Language-Image Pre-training (CLIP)}: CLIP comprises two encoders designed for processing text and visual inputs simultaneously. The visual encoder denoted as $\mathcal{F}_{\theta_v}$, is responsible for mapping visual input $X$ to a fixed-length representation $\bm{f}_v$. On the other hand, the text encoder $\mathcal{F}_{\theta_t}$, is dedicated to text input. The class labels ${y}$ are processed by the text encoder generating latent textual feature denoted as $\bm{f_t}$. The pre-trained parameters for CLIP, represented as ${\theta}_{\mathtt{CLIP}} = \{\theta_{v}, \theta_{t} \}$, are associated with the respective encoders. 
Both the encoders process the input through a sequence of $L$ transformer blocks to produce a latent feature representation.
For zero-shot inference, each text feature with class labels $y \in {1, 2, \cdots, C}$ is paired with the image feature.
The prediction probability on ${X}$ can be expressed as $ p(y_i|{X}) = \frac{\mathtt{exp}(\tau s_i)}{\sum_{j=1}^{C}\mathtt{exp}(\tau s_j)}$, where $ s$ denotes the cosine similarity and $\tau$ represents the softmax temperature parameter.

\vspace{0.10cm}

\noindent\textbf{LoRA Adaptation}: Low Rank Adaptation (LoRA) \cite{hu2021lora} enables parameter efficient training by freezing the model weights and integration of trainable rank decomposition matrices into each layer of the transformer architecture. This results in a substantial reduction in the number of trainable parameters for the downstream adaptation.
We denote the pretrained query $W_Q$, key  $W_K$, and value  $W_V$ projection matrices in the self-attention module jointly by $W$ such that  $W = \{W_Q, W_K, W_V\}$ and ${\Delta W}$ signifies its accumulated gradient update during adaptation. 
Assuming a low intrinsic rank, the pre-trained attention weight matrix $W \in \mathbb{R}^{d\times k}$ undergoes a low-rank decomposition, expressed as $W + {\Delta W} = W + {BA}$, where ${B} \in \mathbb{R}^{d\times r}$, ${A} \in \mathbb{R}^{r\times k}$, and the rank $r \ll min(d, k)$.
During inference, $W$ is frozen and does not receive gradient updates, while $A$ and $B$ contain trainable parameters.
The modified forward pass for any arbitrary input $h$ across the attention module yields $\Tilde{h}$ such that,
\begin{equation}
    \Tilde{h} = Wh + \gamma \cdot{\Delta W}h= Wh + \gamma \cdot{BA}h,
    \label{eqn:lora_standard}
\end{equation}
where $\gamma = \frac{r}{\alpha}$ and $\alpha$ is the scaling factor.

\begin{figure*}[t]
    \centering
    \includegraphics[width=0.85\textwidth]{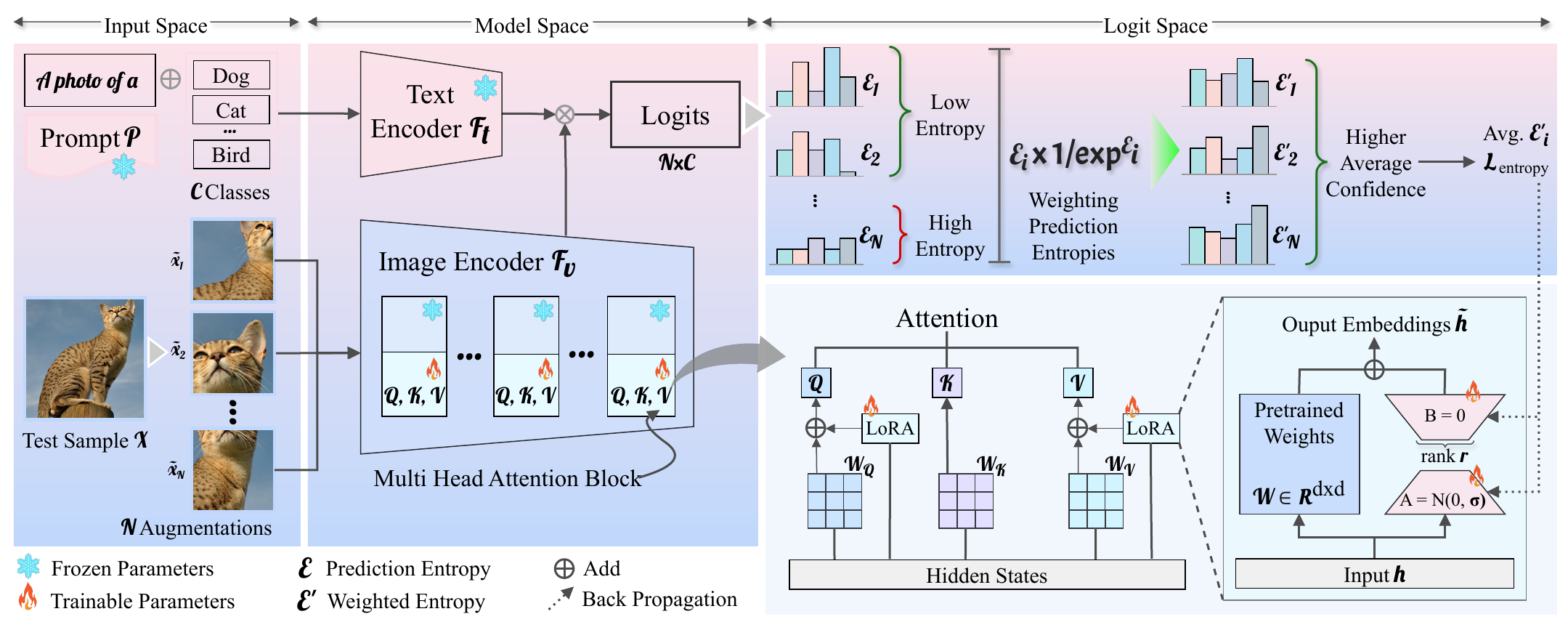}
    \caption{\textbf{Working of Test-Time Low-Rank Adaptation (TTL)}. We integrate parameter efficient low rank matrices into the self-attention module of the image encoder. We adapt these low rank weights on the fly given a single test sample, without the need for pre-trained weights or source data. Maximizing confidence via weighted entropy minimization, TTL updates the low rank weights to optimize the VLM to adapt a test sample in a single update step.}
    \label{fig:method}
\vspace{-0.4cm}
\end{figure*}

\subsection{Proposed Approach: TTL} 
\noindent \textbf{Overview}:
Although current methods \cite{shu2022test,hassan2024align} for prompt-tuning during test time have demonstrated notable success in enhancing CLIP adaptation, the optimal choice between prompt-tuning and alternative approaches remains \textit{unexplored}.
The existing test-time adaptation schemes, as exemplified by \cite{shu2022test} and \cite{hassan2024align}, focus on optimizing prompts for each test sample during inference through entropy minimization. While effective, these approaches have certain limitations. 1). Adaptation using standard entropy minimization is sub-optimal at test time \cite{lee2023entropy}, 2). Prompt tuning is challenging to optimize and 
its performance exhibits non-monotonic changes in trainable parameters, as observed by \cite{hu2021lora} and 3). Some works \cite{hassan2024align} necessitate access to pre-trained prompt weights and source data statistics, which may not be practical at the test-time scenario.
As a solution, illustrated in Figure \ref{fig:method}, we propose integrating LoRA (Low-Rank Adaptation) parameters directly inside the CLIP's visual encoder model to account for the domain shift due to out-of-distribution test sample. As opposed to prompts, LoRA parameters are easier to optimize \cite{hu2021lora} and do not require \textit{pre-trained weights} for initialization or \textit{source data} for alignment, resulting in improved generalization. 
For confidence maximization, we employ weighted entropy loss, as opposed to the standard self-entropy used by \cite{shu2022test,hassan2024align}. The proposed weighted entropy objective results in overall higher average prediction confidence which is beneficial for optimal parameter update, resulting in better prediction accuracy.
Empirical evidence supporting the advantages of weighted entropy loss over standard self-entropy is presented in Table \ref{tab:weighted_entropy}.

\vspace{0.10cm}

\noindent{\textbf{Low-Rank Adaptation at Test-Time}}:
\label{Low-Rank Adaptation at Test-Time}
For parameter efficient adaptation at test-time, we integrate LoRA parameters inside the attention layers of the CLIP's visual encoder.
As indicated by \cite{hu2021lora}, over-parameterized models exhibit low intrinsic dimension, and the change in weights during model adaptation has a \textit{low intrinsic rank}. We extend this hypothesis to test-time adaptation, where only a single test sample is available, implying that updating only a few parameters is sufficient for efficient and effective adaptation.
Given a test sample $X \in \mathcal{D}^{test}$, we take $N$ randomly augmented views using transformation function $\mathcal{H}$ such that we get a batch of images denoted as  $\mathcal{H}(X)$. Low rank weight matrices are introduced in the query ($W_{Q}$) and value  ($W_{V}$) projection layers inside the self-attention module and are jointly parameterised by $\Phi$.
Let $h^{l}_*$ be the input features to the self-attention module of $l^{th}$ encoder block and $\Tilde{h}^{l}_*$ be the corresponding output, then the forward pass (Eq. \ref{eqn:lora_standard}) can be expressed as,
\begin{equation}
    \Tilde h^l_{Q} = W^l_{Q} h^l_{Q} + \gamma \cdot\Delta{W_{\Phi;Q}^l}h^l_{Q} =  W^l_{Q}h^l_{Q} + \gamma\cdot({B^{l}_{Q} A^{l}_{Q}})_\Phi h^l_{Q} \nonumber
\end{equation}
\begin{equation}
    \Tilde h^l_{V} = W^l_{V} h^l_{V} + \gamma \cdot\Delta{W_{\Phi;V}^l}h^l_{V} =  W^l_{V}h^l_{V} + \gamma\cdot({B^{l}_{V} A^{l}_{V}})_\Phi h^l_{V} \nonumber
\end{equation}
 At the test-time, we optimize the rank decomposition matrices corresponding to query and value projection matrices in the self-attention module while keeping the pre-trained weights of CLIP frozen. In general, the optimization objective for a randomly augmented view $\Tilde{x} \in \mathcal{H}(X)$ can be constructed as,
 \begin{equation}
        \Phi^{*} =  \text{arg}\min_{\Phi} \mathcal{L}(\mathcal{F}_{\theta_{\mathtt{CLIP}}},\Phi, \Tilde{x})
\end{equation}
Since LoRA parameters directly influence the model attention, it leads to better predictions by concentrating the model attention on the object of interest (see Appendix \ref{appendix:attn}). Standard optimization objective such as self-entropy loss gives equal weight to all the augmented views leading towards sub-optimal optimization, while as confidence selection ignores certain views which may be beneficial for the correct prediction. To this end, we propose to update the parameters with weighted entropy objective which gives variable weight to each view.

\vspace{0.10cm}

\noindent{\textbf{Weighted Entropy Minimization}}:
Instead of discarding majority of crops of test sample based on confidence selection as done in previous works \cite{shu2022test, hassan2024align}, we take a slightly different approach and utilize all the crops for optimization. 
As discussed in Sec. \ref{sec:intro}, relying solely on entropy for confidence selection is not consistently reliable due to model focusing on unwanted elements in the input. 
Our analysis in Figure \ref{fig:entropy_octiles} reinforces this observation, highlighting a weaker correlation between confidence selection and the true model prediction during test-time. This signifies that predictions from augmented views in the highest confidence quartile may not consistently contribute favorably to model updates (Figure \ref{fig:entropy_octiles}) as further validated by \cite{lee2023entropy}.
Therefore, we introduce weighted entropy loss which  encompasses all the augmented views at test-time and assigns variable weights to model's prediction of each view resulting in overall higher average confidence. 
For each augmented view $\Tilde{x} \in \mathcal{H}(X)$, we map its visual features to the class labels and compute a standard self-entropy loss across $L$ encoder blocks represented as,
\vspace{-0.2cm}
\begin{align}
\label{eq:tpt_3}
    \mathcal{L}_{\Phi}(\Tilde{x})= \sum_{i=1}^{C}\Tilde{p}_{\Phi} (y_i|\Tilde{x}) \log \Tilde{p}_{\Phi}(y_i|\Tilde{x}),
\end{align}
where $\Tilde{p}_\Phi(y_i|\Tilde{x})$ represents the vector class probabilities produced by the model.
The final objective is thus the weighted sum of the individual entropy losses corresponding to each augmented view $\Tilde{x}$. The final objective function for parameter update is given as,
\begin{align}
\label{eq:tpt_1}
    \text{arg}&\min_{\Phi}- \frac{1}{N}\sum_{\Tilde{x} \in \mathcal{H}(X)} \beta_{\Phi}(\Tilde{x}) \cdot \mathcal{L}_{\Phi}(\Tilde{x})
\end{align}
Here $\beta_{\Phi}(\Tilde{x})$ is the weight coefficient of the augmented view corresponding to $\Tilde{x}$ and is expressed as,
\begin{equation}
\label{eq:tpt_2}
    \beta_{\Phi}(\Tilde{x}) = \frac{1}{\exp(\mathcal{L}_{\Phi}(\Tilde{x}) - \varepsilon)},
\end{equation}
where $\varepsilon$ is a normalization factor.
The resulting objective function in Eq. \ref{eq:tpt_1} maximizes the average confidence and enhances the model's prediction of test sample.

\section{Experiments and Results}

\subsection{Experimental Setup}
\label{sec:exp}
\noindent\textbf{Implementation Details}:
We initialize trainable LoRA matrices with random $Xavier$ initialization \cite{glorot2010understanding} and set rank $r=16$ and $\alpha=32$. Optimization of the LoRA weights occurs in layers 10 to 12 within the vision branch, involving a single-step update using a single test sample. We obtain 63 augmented views of input sample using random resized crops
and horizontal flip augmentations to construct a batch of 64 images including the original image
to mimic the setting of TPT. We utilize all the 64 crops to compute the average prediction probability and optimize the LoRA parameters to minimize
the weighted version of combined average prediction entropy loss using the AdamW optimizer. We use a learning rate of 5e-3 for all the cases and set the normalization constant $\mathcal{E}$ equal to 0.4. A fixed prompt template "a photo of a" is used with the classnames. The entire setup runs on a single NVIDIA A100 40GB GPU. 
\vspace{0.10cm}

\noindent\textbf{Datasets}:
We assess the performance of our proposed method in two cases \ie \texttt{Natural Distribution Shifts} ($\mathcal{C}_1$) and \texttt{Cross-Datasets Generalization}  ($\mathcal{C}_2$) in accordance with ~\cite{shu2022test, hassan2024align}. For  $\mathcal{C}_1$, we utilize four datasets—ImageNet-V2 ~\cite{recht2019imagenet}, {ImageNet-A~\cite{hendrycks2021natural}, ImageNet-R~\cite{hendrycks2021many}, and ImageNet-Sketch~\cite{wang2019learning}—as out-of-distribution (OOD) data for ImageNet~\cite{deng2009imagenet}, to assess the effectiveness of our method. For  $\mathcal{C}_2$, we utilize 10 image classification datasets covering a diverse range of visual recognition tasks. This set includes the generic-objects dataset Caltech101~\cite{fei2004learning} and five fine-grained datasets: OxfordPets~\cite{parkhi2012cats}, StanfordCars~\cite{krause20133d}, Flower102~\cite{nilsback2008automated}, Food101~\cite{bossard2014food}, and FGVC-Aircraft~\cite{maji2013fine}. These fine-grained datasets encompass images of animals, flowers, and transportation. Additionally, four datasets covering scenes, textures, satellite imagery, and human actions are considered: SUN397~\cite{xiao2010sun}, DTD~\cite{cimpoi2014describing}, EuroSAT~\cite{helber2019eurosat}, and UCF101~\cite{soomro2012dataset}.

{\renewcommand{\arraystretch}{1.0}
\begin{table*}[t]
\caption{\textbf{Top 1 accuracy} $\%$ of state-of-the-art baselines under \texttt{strict zero-shot settings}, where \textbf{ImageNet-Sk.} indicates the ImageNet-Sketch dataset, \textbf{OOD Avg.} indicates the OOD average results. {\cellcolor{mygray}{$bs.$}} indicates the baseline, \ie,CLIP-ViT-B-16. The arrow ${\color{myforestgreen}\uparrow}$ and ${\color{red}\downarrow}$ indicate \textbf{improvements} and \textbf{decrements} compared with {\cellcolor{black}{$bs.$}}. Detailed analyses are provided in Sec.~\ref{sec:results}.}
\centering
\resizebox{\textwidth}{!}{%
\begin{tabular}{l|ccccc|c|c}
\hline
\textbf{Method} & \textbf{ImageNet} & \textbf{ImageNet-A} & \textbf{ImageNet-V2} & \textbf{ImageNet-R} & \textbf{ImageNet-Sk.} & \textbf{Average} & \textbf{OOD Avg.} \\ 
\cmidrule(r){1-1} \cmidrule(lr){2-2} \cmidrule(lr){3-3} \cmidrule(lr){4-4} \cmidrule(lr){5-5} \cmidrule(lr){6-6} \cmidrule(lr){7-7} \cmidrule(l){8-8}

\rowcolor{mygray} 
CLIP-ViT-B/16 & 67.30\stdvuno{$bs.$} & 47.14\stdvuno{$bs.$} & 59.90\stdvuno{$bs.$} & 71.20\stdvuno{$bs.$} & 43.00\stdvuno{$bs.$} & 57.71\stdvuno{$bs.$} & 55.31\stdvuno{$bs.$} \\

\rowcolor[HTML]{f6fbff} 
Ensemble & 68.50\stdvun{1.20} & 48.44\stdvun{1.30} & 62.70\stdvun{2.80} & 73.50\stdvun{2.30} & 45.50\stdvun{2.50} & 59.73\stdvun{2.02} & 57.53\stdvun{2.22} \\

\rowcolor[HTML]{f6fbff} 
CoOp${\color{cyan}_{2021}}$ & \textbf{72.30}\stdvun{5.00} & 49.25\stdvun{2.11} & 65.70\stdvun{5.80} & 71.50\stdvun{0.30} & 47.60\stdvun{4.60} & 61.27\stdvun{3.56} & 58.51\stdvun{3.20} \\

\rowcolor[HTML]{f6fbff} 
CoCoOp${\color{cyan}_{2022}}$ & 71.40\stdvun{4.10} & 50.05\stdvun{2.91} & 63.80\stdvun{3.90} & 73.10\stdvun{1.90} & 46.70\stdvun{3.70} & 61.01\stdvun{3.30} & 58.41\stdvun{3.10} \\

\rowcolor[HTML]{eff6fc} 
TPT${\color{cyan}_{2022}}$ & 68.90\stdvun{1.60} & 54.59\stdvun{7.45} & 63.13\stdvun{3.23} & 77.05\stdvun{5.85} & 47.99\stdvun{4.99} & 62.33\stdvun{4.62} & 60.69\stdvun{5.38} \\

\rowcolor[HTML]{e7f3fa}
CALIP${\color{cyan}_{2023}}$ & 66.74\stdvuru{0.56} & 47.76\stdvun{0.62} & 60.76\stdvun{0.86} & 73.99\stdvun{2.79} & 46.12\stdvun{3.12} & 59.07\stdvun{1.36} & 57.16\stdvun{1.85} \\

\rowcolor[HTML]{e3f2ff} 
PromptAlign${\color{cyan}_{2023}}$ & 60.02\stdvuru{7.28} & 45.52\stdvuru{1.62} & 54.53\stdvuru{5.37} & 72.84\stdvun{1.64} & 37.72\stdvuru{5.28} & 54.13\stdvuru{3.58} & 52.65\stdvuru{2.66} \\

\rowcolor[HTML]{deefff} 
\textbf{TTL (Ours)} & 70.23\stdvun{2.93} & \textbf{60.51}\stdvun{13.37} & \textbf{64.55}\stdvun{4.65} & \textbf{77.54}\stdvun{6.34} & \textbf{48.61}\stdvun{5.61} & \textbf{64.29}\stdvun{6.58} & \textbf{62.80}\stdvun{7.49} \\ \hline

\end{tabular}
}
\label{tab:natural_shift}
\end{table*}

{\renewcommand{\arraystretch}{1.0}
\begin{table*}[t]

\caption{\textbf{Top 1 accuracy} $\%$ of state-of-the-art baselines under \texttt{strict zero-shot settings}, where \textbf{Average} indicates average accuracies of the \texttt{Cross-Datasets Generalization}. The arrow ${\color{myforestgreen}\uparrow}$ and ${\color{red}\downarrow}$ indicate \textbf{improvements} and \textbf{decrements} of our method against the CLIP method, \ie, CLIP-ViT-B/16. Detailed analyses are provided in Sec.~\ref{sec:results}.}

\centering
\resizebox{\textwidth}{!}{%
\begin{tabular}{l|cccccc} \hline
\textbf{Method} & ~\textbf{Flower102}\cite{nilsback2008automated}~ & ~\textbf{DTD}\cite{cimpoi2014describing}~ & ~\textbf{OxfordPets}\cite{parkhi2012cats}~ & ~\textbf{UCF}\cite{soomro2012dataset}~ & ~\textbf{Caltech101}\cite{fei2004learning}~ & ~\textbf{Aircraft}\cite{maji2013fine}~ \\ 
\cmidrule(r){1-1} \cmidrule(lr){2-2} \cmidrule(r){3-3} \cmidrule(r){4-4} \cmidrule(r){5-5} \cmidrule(l){6-6} \cmidrule(l){7-7}

\rowcolor{mygray} 
CLIP-ViT-B/16 & 67.94\stdvuno{$bs.$} & 44.10\stdvuno{$bs.$} & 85.71\stdvuno{$bs.$} & 63.37\stdvuno{$bs.$} & 90.29\stdvuno{$bs.$} & 24.70\stdvuno{$bs.$} \\

\rowcolor[HTML]{F6FBFF} 
Ensemble & 67.65 & 44.87 & 86.20 & 64.36 & 90.89 & 24.40 \\

\rowcolor[HTML]{F6FBFF} 
CoOp${\color{cyan}_{2021}}$ \cite{zhou2022coop} & 66.08 & 42.17 & \textbf{89.00} & 66.04 & 91.69 & 18.00 \\

\rowcolor[HTML]{f6fbff} 
CoCoOp${\color{cyan}_{2022}}$ \cite{zhou2022cocoop} & 70.88 & 44.78 & 88.71 & 68.42 & 92.49 & 24.20 \\

\rowcolor[HTML]{eff6fc} 
TPT${\color{cyan}_{2022}}$ \cite{shu2022test} & 69.31\stdvun{1.37} & 46.23\stdvun{2.13} & 86.49\stdvun{0.78} & 66.44\stdvun{3.07} & 92.49\stdvun{2.20} & \textbf{24.90}\stdvun{0.20} \\

\rowcolor[HTML]{e7f3fa} 
CALIP${\color{cyan}_{2023}}$ \cite{guo2023calip} & 67.64\stdvuru{0.3} & 44.44\stdvun{0.34} & 87.82\stdvun{2.11} & 64.05\stdvun{0.68} & 93.27\stdvun{2.98} & 24.12\stdvuru{0.58} \\

\rowcolor[HTML]{e3f2ff} 
PromptAlign${\color{cyan}_{2023}}$ \cite{hassan2024align} & 51.60\stdvuru{16.34} & 27.60\stdvuru{16.50} & 75.82\stdvuru{9.89} & 57.31\stdvuru{6.06} & 87.18\stdvuru{3.11} & 6.96\stdvuru{17.74} \\

\rowcolor[HTML]{deefff} 
\textbf{TTL (Ours)} & \textbf{70.48}\stdvun{2.54} & \textbf{46.69}\stdvun{2.59} & 88.72\stdvun{3.01} & \textbf{69.20}\stdvun{5.83} & \textbf{93.63}\stdvun{3.34} & 23.82\stdvuru{1.78} \\ \hline

\end{tabular}
}

\centering

\resizebox{\textwidth}{!}{%
\begin{tabular}{l|cccc|c} \hline

\textbf{Method} & ~~~\textbf{EuroSAT}\cite{helber2019eurosat}~~~ & ~~~\textbf{StanfordCars}\cite{krause20133d}~~~ & ~~~\textbf{Food101}\cite{bossard2014food}~~~ & ~~~\textbf{SUN397}\cite{xiao2010sun}~~~ & ~~~~~\textbf{Average}~~~~~ \\ 

\cmidrule(r){1-1} \cmidrule(lr){2-2} \cmidrule(r){3-3} \cmidrule(r){4-4} \cmidrule(r){5-5} \cmidrule(l){6-6} 

\rowcolor{mygray} 
CLIP-ViT-B/16 & 40.64\stdvuno{$bs.$} & 66.58\stdvuno{$bs.$} & 85.05\stdvuno{$bs.$} & 61.88\stdvuno{$bs.$} & 63.03\stdvuno{$bs.$} \\

\rowcolor[HTML]{F6FBFF} 
Ensemble & 47.01 & 67.60 & 85.35 & 64.65 & 64.30 \\

\rowcolor[HTML]{F6FBFF} 
CoOp${\color{cyan}_{2021}}$ \cite{zhou2022coop} & 35.36 & 63.44 & 85.15 & 61.54 & 61.85 \\

\rowcolor[HTML]{F6FBFF} 
CoCoOp${\color{cyan}_{2022}}$ \cite{zhou2022cocoop} & 39.23 & 65.22 & 86.53 & 64.65 & 64.51 \\

\rowcolor[HTML]{eff6fc} 
TPT${\color{cyan}_{2022}}$ \cite{shu2022test} & 37.15\stdvuru{3.49} & 66.50\stdvuru{0.08} & \textbf{86.93}\stdvun{1.88} & 63.48\stdvun{1.60} & 63.99\stdvun{0.96} \\

\rowcolor[HTML]{e7f3fa} 
CALIP${\color{cyan}_{2023}}$ \cite{guo2023calip} & \textbf{42.27}\stdvun{1.63} & 65.80\stdvuru{0.78} & 82.76\stdvuru{2.29} & 62.52\stdvun{0.64} & 63.47\stdvun{0.44} \\

\rowcolor[HTML]{e3f2ff} 
PromptAlign${\color{cyan}_{2023}}$ \cite{hassan2024align} & 35.57\stdvuru{5.07} & 58.70\stdvuru{7.88} & 82.23\stdvuru{2.82} & 57.84\stdvuru{4.04} & 54.08\stdvuru{8.95} \\

\rowcolor[HTML]{deefff} 
\textbf{TTL (Ours)} & 42.02\stdvun{1.38} & \textbf{67.96}\stdvun{1.38} & 85.05\stdvueq{0.00} & \textbf{66.32}\stdvun{4.44} & \textbf{65.39}\stdvun{2.36} \\ \hline

\end{tabular}
}
\label{tab:cross_domain}
\end{table*}

\vspace{0.10cm}
\noindent\textbf{Baselines}:
To evaluate our proposed approach, we adopt two groups of VLM methods: $\mathcal{M}_1$, which perform strict zero-shot classification, \ie without any form of few-shot pre-training or external model support, for fair comparison; and $\mathcal{M}_2$, standard baselines followed in the recent state-of-the-art methods.
\vspace{-0.20cm}
\begin{itemize}[left=0pt]
    \item For $\mathcal{M}_1$: TPT \cite{shu2022test}, a state-of-the-art test-time prompt tuning method optimizing learnable prompts across multiple augmented views;
    PromptAlign \cite{hassan2024align}, an extension of TPT that incorporates multi-modal prompt learning for explicit alignment of feature distributions;
    CALIP \cite{guo2023calip}, introduces parameter-free attention to enhance the exchange of informative features between images and text in CLIP;
    and standard zero-shot CLIP is included with default configurations.
\vspace{-0.20cm}
    \item For $\mathcal{M}_2$:
    CoOp \cite{zhou2022coop}, a few-shot prompt tuning method that adjusts a template prompt for each downstream task;
    CoCoOp \cite{zhou2022cocoop}: an enhanced method for few-shot prompt-tuning that generates input-conditional prompts using a lightweight neural network;
    and zero-shot CLIP with an Ensemble \cite{radford2021learning} of 80 specially crafted prompts.
\end{itemize}

\vspace{0.05cm}
\noindent\textbf{Reproducibility}: All baselines are reproduced on our system to ensure fairness. PromptAlign \cite{hassan2024align} is implemented without pre-trained weights. DiffTPT \cite{feng2023diverse} is not considered due to impracticality in replicating the method, given the extensive time required for generating diffusion-based samples during inference. Additionally, their reported scores with 4 update steps would not offer a fair comparison.

\subsection{Main Results}
\label{sec:results}

\noindent\textbf{Natural Distribution Shifts}:
Table \ref{tab:natural_shift} summarizes the evaluation of our method comparing $\mathcal{M}_1$ and $\mathcal{M}_2$ baseline methods under $\mathcal{C}$ase 1 with ViT-B/16 backbone under \texttt{strict zero-shot} settings. We can see that:
\noindent\textbf{Our approach outperforms all four $\mathcal{M}_1$ and three $\mathcal{M}_2$ baseline methods}, across all four out-of-distribution (OOD) datasets, including the in-domain ImageNet, demonstrating a substantial increase in generalization performance. Specifically, across the in-domain dataset, TTL exhibits significantly improved generalization accuracy compared to CLIP, TPT, PromptAlign, CALIP ($\mathcal{M}_1$ methods), Ensemble, CoOp, and CoCoOp ($\mathcal{M}_2$ methods) increasing average accuracy from 57.71, 62.33, 54.13, 59.07, 59.73, 61.27, and 61.01 to \textbf{64.29}, respectively. Similarly, across the average OOD dataset, TTL shows consistent performance gain in handling natural distribution shifts compared to CLIP, TPT, PromptAlign, CALIP ($\mathcal{M}_1$ methods), Ensemble, CoOp, and CoCoOp ($\mathcal{M}_2$ methods) improving from 55.31, 60.69, 52.65, 57.16, 57.53, 58.51, and 58.41 to \textbf{62.80}, respectively. 
Supporting our initial hypothesis that low-rank attention weight adaptation for a single test sample improves the generalization, this highlights the superior adaptability of our method in handling OOD domain shifts. This establishes TTL as a \textit{go-to} choice over SOTA $\mathcal{M}_1$ methods like TPT and PromptAlign, which update learnable prompts during inference.

\noindent\textbf{Generalization to Cross-Dataset Transfer}:
To evaluate the generalization performance of our proposed method and baselines on the 10 $\mathcal{C}_2$ datasets, we analyze the results within the \texttt{strict zero-shot} settings, as shown in Table \ref{tab:cross_domain}. We can see that: 
\textbf{Our method outperforms all seven baselines on average across all $\mathcal{C}_2$ datasets}, individually surpassing six out of ten $\mathcal{C}_2$ datasets. Among the four $\mathcal{M}_1$ methods, TTL exhibits consistent improvements, outperforming CLIP, TPT, CALIP, and PromptAlign with average improvements of +2.36, +1.40, +1.91, and +11.31, reaching up to \textbf{65.39} average accuracy. Additionally, compared to three $\mathcal{M}_2$ methods, TTL achieves average improvements of +1.09, +3.53, and +0.88 when compared to Ensemble, CoOp, and CoCoOp, respectively. 
These results affirm that our proposed TTL, which combines the effect of weighted entropy minimization and low-rank adaptation, achieves superior distribution alignment compared to all baselines, including CLIP. This indicates that our method is a promising approach for zero-shot adaptation and demonstrates robustness to cross-data distributional variations.


{\renewcommand{\arraystretch}{1.0}
\begin{table*}[t]

\caption{\textbf{Effect of Weighted Entropy} under \texttt{strict zero-shot settings}, where \textbf{Average} indicates average accuracies of the \texttt{Cross-Datasets Generalization}. 'TTL \textit{w/o} Wt. Ent.' indicate TTL without weighted entropy approach (See Table \ref{tab:weighted_entropy_rebut}). 
}

\centering
\resizebox{\textwidth}{!}{%
\begin{tabular}{l|cccccc} \hline
Method & ~\textbf{Flower102}\cite{nilsback2008automated}~ & ~\textbf{DTD}\cite{cimpoi2014describing}~ & ~\textbf{OxfordPets}\cite{parkhi2012cats}~ & ~\textbf{UCF}\cite{soomro2012dataset}~ & ~\textbf{Caltech101}\cite{fei2004learning}~ & ~\textbf{Aircraft}\cite{maji2013fine}~ \\ 
\cmidrule(r){1-1} \cmidrule(lr){2-2} \cmidrule(r){3-3} \cmidrule(r){4-4} \cmidrule(r){5-5} \cmidrule(l){6-6} \cmidrule(l){7-7}

\rowcolor{mygray} 
CLIP-ViT-B/16 & 67.94\stdvuno{$bs.$} & 44.10\stdvuno{$bs.$} & 85.71\stdvuno{$bs.$} & 63.37\stdvuno{$bs.$} & 90.29\stdvuno{$bs.$} & 24.70\stdvuno{$bs.$} \\

\rowcolor[HTML]{eff6fc} 
TPT${\color{cyan}_{2022}}$ \cite{shu2022test} & 69.31\stdvun{1.37} & 46.23\stdvun{2.13} & 86.49\stdvun{0.78} & 66.44\stdvun{3.07} & 92.49\stdvun{2.20} & \textbf{24.90}\stdvun{0.20} \\

\rowcolor[HTML]{e7f3fa} 
TPT \textit{w} Wt. Ent. & 69.56\stdvun{1.62} & 46.69\stdvun{2.59} & 88.58\stdvun{2.87} & 69.18\stdvun{5.81} & 93.55\stdvun{3.26} & 23.14\stdvuru{1.56} \\

\rowcolor[HTML]{e7f3fa} 
TTL \textit{w/o} Wt. Ent. & 68.78\stdvun{0.84} & 45.57\stdvun{1.47} & \textbf{88.91}\stdvun{3.20} & 68.09\stdvun{4.72} & \textbf{94.04}\stdvun{3.75} & 24.72\stdvuru{0.88} \\

\rowcolor[HTML]{deefff} 
\textbf{TTL (Ours)} & \textbf{70.48}\stdvun{2.54} & \textbf{46.69}\stdvun{2.59} & 88.72\stdvun{3.01} & \textbf{69.20}\stdvun{5.83} & 93.63\stdvun{3.34} & 23.82\stdvuru{1.78} \\ \hline

\end{tabular}
}

\centering

\resizebox{\textwidth}{!}{%
\begin{tabular}{l|cccc|c} \hline

Method & ~~~\textbf{EuroSAT}\cite{helber2019eurosat}~~~ & ~~~\textbf{StanfordCars}\cite{krause20133d}~~~ & ~~~\textbf{Food101}\cite{bossard2014food}~~~ & ~~~\textbf{SUN397}\cite{xiao2010sun}~~~ & ~~~~~\textbf{Average}~~~~~ \\ 

\cmidrule(r){1-1} \cmidrule(lr){2-2} \cmidrule(r){3-3} \cmidrule(r){4-4} \cmidrule(r){5-5} \cmidrule(l){6-6} 

\rowcolor{mygray} 
CLIP-ViT-B/16 & 40.64\stdvuno{$bs.$} & 66.58\stdvuno{$bs.$} & 85.05\stdvuno{$bs.$} & 61.88\stdvuno{$bs.$} & 63.03\stdvuno{$bs.$} \\

\rowcolor[HTML]{eff6fc} 
TPT${\color{cyan}_{2022}}$ \cite{shu2022test} & 37.15\stdvuru{3.49} & 66.50\stdvuru{0.08} & \textbf{86.93}\stdvun{1.88} & 63.48\stdvun{1.60} & 63.99\stdvun{0.96} \\

\rowcolor[HTML]{e7f3fa} 
TPT \textit{w} Wt. Ent. & 41.96\stdvun{1.32} & 66.37\stdvuru{0.21} & 84.92\stdvuru{0.13} & 64.96\stdvun{3.08} & 64.89\stdvun{1.86} \\

\rowcolor[HTML]{e7f3fa} 
TTL \textit{w/o} Wt. Ent. & \textbf{42.07}\stdvun{1.43} & 66.75\stdvun{0.17} & 83.65\stdvuru{1.40} & 62.59\stdvun{0.71} & 64.52\stdvun{1.40} \\

\rowcolor[HTML]{deefff} 
\textbf{TTL (Ours)} & 42.02\stdvun{1.38} & \textbf{67.96}\stdvun{1.38} & 85.05\stdvueq{0.00} & \textbf{66.32}\stdvun{4.44} & \textbf{65.39}\stdvun{2.36} \\ \hline

\end{tabular}
}
\label{tab:weighted_entropy}
\end{table*}

\noindent\textbf{Low-Rank Adaptation \textit{vs.} Prompt Tuning}:
We assess TTL's performance against various prompt tuning approaches in \texttt{strict zero-shot} scenarios, categorizing these approaches into Text Prompt $\mathcal{T}_p$, Visual Prompt $\mathcal{V}_p$, and Multi-Modal Prompt $\mathcal{N}_p$.
As illustrated in Figure \ref{fig:maple_compare_short} (and Figure \ref{fig:maple_compare} in Appendix), text prompt-tuning approaches $\mathcal{T}_p$ like TPT \cite{shu2022test} is effective for test-time zero-shot adaptation as it learns a text prompt instead of a standard template. However, approaches utilizing visual $\mathcal{V}_p$ and multi-modal prompts $\mathcal{N}_p$ at test-time without pre-training not only fail to improve but also show reduced generalization compared to base CLIP across diverse domain shifts.

Supporting our observation, the $\mathcal{N}_p$ method PromptAlign \cite{hassan2024align}, which appends prompts to both text and image encoders, shows that explicitly aligning visual feature distributions with an alignment loss enhances CLIP's generalization. However, this effectiveness relies on few-shot pre-trained multi-modal prompts for test-time adaptation. Without pre-training, PromptAlign underperforms base CLIP. When pre-trained, PromptAlign surpass zero-shot methods, but at the expense of eliminating the essence of zero-shot generalization, essential for real-world scenarios (See Table \ref{tab:tick_methods}). This reliance on pre-trained prompts indicates that without embedded prior knowledge, PromptAlign struggles with distribution shifts and meaningful representation learning. In contrast, TTL outperforms all $\mathcal{T}_p$, $\mathcal{V}_p$, and $\mathcal{N}_p$ approaches, achieving higher performance with respective gains of +2.11, +8.98, and +10.00 in $\mathcal{C}_1$ and $\mathcal{C}_2$ cases without any pre-training (Figure \ref{fig:tsne_small}).

\begin{figure}[t]
    \centering
    \begin{subfigure}[b]{0.235\textwidth}
        \includegraphics[width=\textwidth]{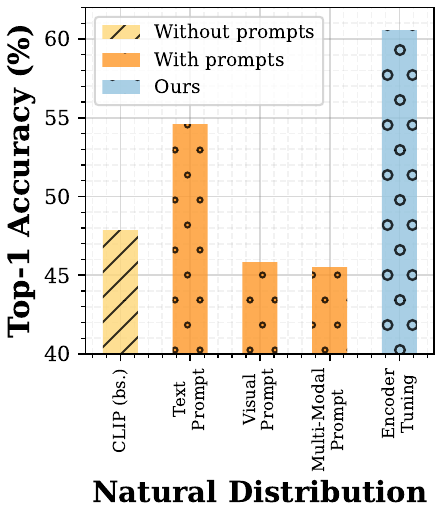}
    \end{subfigure}
    \hfill
    \begin{subfigure}[b]{0.235\textwidth}
        \includegraphics[width=\textwidth]{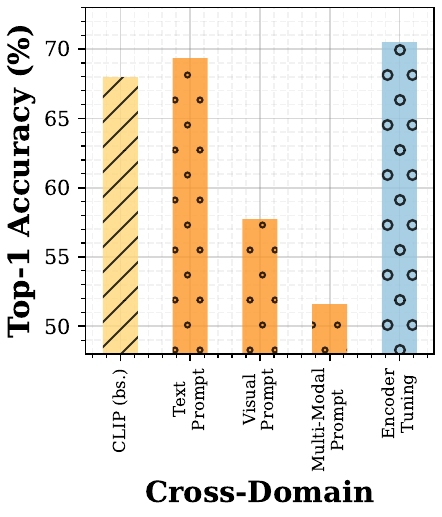}
    \end{subfigure}
    \caption{\textbf{Test-time performance of zero-shot methods.} CLIP \textit{vs.} Textual Prompt Tuning (TPT) \textit{vs.} Visual Prompt Tuning \textit{vs.} Multi-modal Prompt Tuning \textit{vs.} \textbf{TTL (Ours)} (See Figure \ref{fig:maple_compare}). 
    }
    \label{fig:maple_compare_short}
\vspace{-0.4cm}
\end{figure}

\begin{figure}[t]
    \vspace{-0.15cm}

    \centering
    \begin{subfigure}[b]{0.232\textwidth}
        \includegraphics[width=\textwidth]{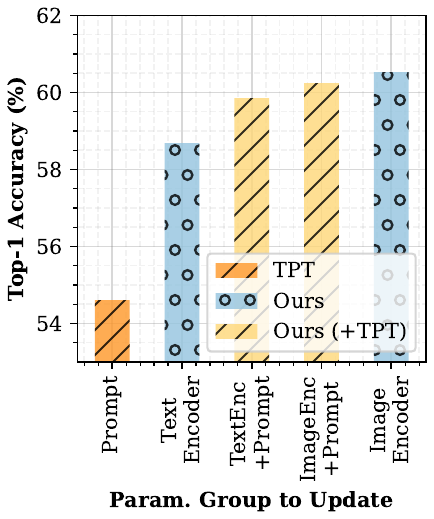}
        \label{fig:param_group}
    \end{subfigure}
    \begin{subfigure}[b]{0.235\textwidth}
        \includegraphics[width=\textwidth]{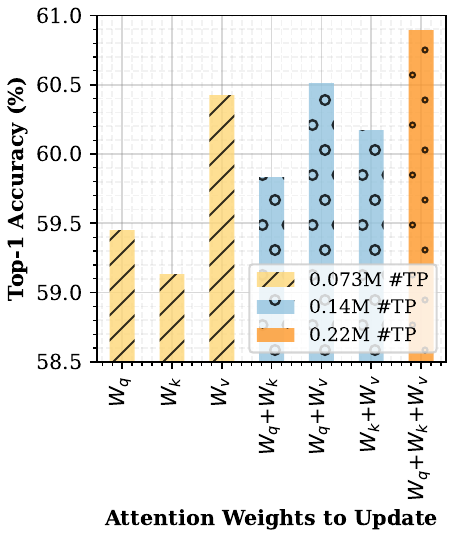}
        \label{fig:attn_group}
    \end{subfigure}
    \vspace{-0.5cm}
    \caption{\textbf{Test-time Low-Rank Adaption} across (a) (left) different combinations of trainable model components (b) (right) different combinations of query, key, and value of image encoder.}
    \vspace{-0.35cm}
    \label{fig:param_attn}
\end{figure}


\section{Analysis and Ablation}
\label{sec:ablation}
We conduct a range of empirical analyses and ablation studies to assess the impact of different design choices in our method. Unless specified otherwise, we present the analyses using the ImageNet-A dataset with ViT-B/16 backbone, opting for the smallest domain generalization variant for simplicity. 

\vspace{0.10cm}
\noindent\textbf{Optimizing Different Parameter Groups}:
We investigate the effectiveness of optimizing different components within TTL framework for test-time adaptation. 
We compare four different parameter groups for optimization at test-time: \texttt{Text Encoder + TTL}, \texttt{Text Encoder + Prompt}, \texttt{Image Encoder + Prompt}, and \texttt{Image Encoder + TTL}.
From Figure \ref{fig:param_attn}{a}, we notice that simply optimizing TTL in the \texttt{Text Encoder} achieves higher performance than prompt tuning methods. Additionally, utilizing TTL inside \texttt{Image Encoder} achieves the maximum performance gain compared to other parameter groups.

\vspace{0.10cm}
\noindent\textbf{Optimizing Different Attention Groups}:
\noindent We investigate the impact of optimizing different combinations of attention weight groups ($W_Q$, $W_K$, and $W_V$) within the self-attention module. The results, as depicted in Figure \ref{fig:param_attn}{b}, indicate that optimizing LoRA parameters in $W_Q$\texttt{+}$W_K$\texttt{+}$W_V$ produces maximum performance gains. However, there is a trade-off as the inclusion of more weight groups results in linear increment of trainable parameters. Therefore, to ensure computation efficiency while maintaining decent performance, we optimize LoRA parameters within $W_Q$ and $W_V$ projection layers.

\vspace{0.10cm}
\noindent\textbf{Analysing the Effect of Weighted Entropy}: We observe that instead of discarding the low entropy augmented views and assigning variable weights to the self-entropy of the predictions from each augmented view is advantageous for test-time optimization.
Specifically, it is important to recognize that \textit{a model predicting a test sample with high confidence \ie low entropy, does not necessarily indicate a correct prediction} (Figure \ref{fig:entropy_octiles}). Table \ref{tab:weighted_entropy} (\& Table \ref{tab:weighted_entropy_rebut} in Appendix) illustrates a notable +0.87 average performance gain achieved by our method when using weighted entropy, surpassing the performance without weighted entropy. This integration effectively considers the contribution of both low and high entropy samples during optimization, enhancing test-time adaptation and improving generalization across both $\mathcal{C}_1$ and $\mathcal{C}_2$ datasets as shown in Table \ref{tab:natural_shift} and \ref{tab:cross_domain}.

\begin{figure}[t!]
    \centering
    \begin{subfigure}[b]{0.23\textwidth}
        \includegraphics[width=\textwidth]{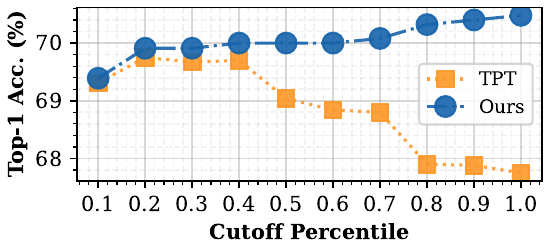}
        \caption{Cutoff Percentile \textit{vs.} Acc.}
        \label{fig:cutoff_percent}
    \end{subfigure}
    \begin{subfigure}[b]{0.23\textwidth}
        \includegraphics[width=\textwidth]{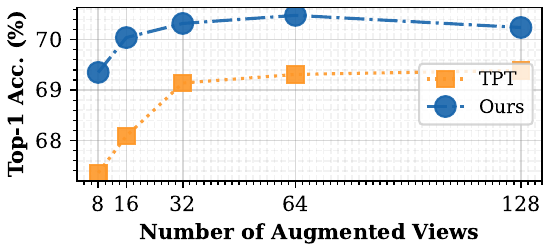}
        \caption{\#Augmentations \textit{vs.} Acc.}
        \label{fig:aug_views}
    \end{subfigure}
    \begin{subfigure}[b]{0.23\textwidth}
        \includegraphics[width=\textwidth]{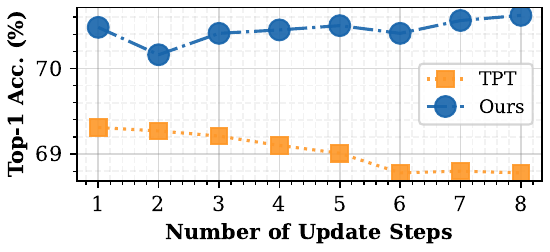}
        \caption{\#Update steps \textit{vs.} Accuracy}
        \label{fig:update_steps}
    \end{subfigure}
    \caption{Analysis of compute resource constraints on performance across Cross-domain data on average.}
    \label{fig:cutoff_augs_steps}
\vspace{-0.45cm}
\end{figure}

\begin{figure}[h]
    \centering    
    \begin{subfigure}[b]{0.15\textwidth}
        \includegraphics[width=\textwidth]{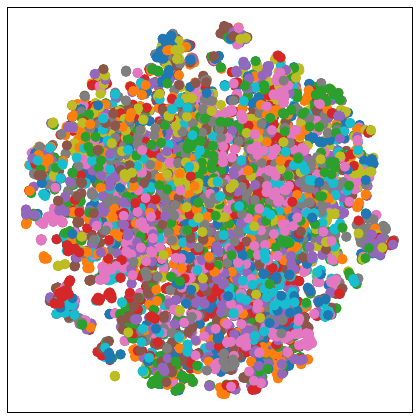}
        \caption{TPT \cite{shu2022test}}
    \end{subfigure}
    \begin{subfigure}[b]{0.15\textwidth}
        \includegraphics[width=\textwidth]{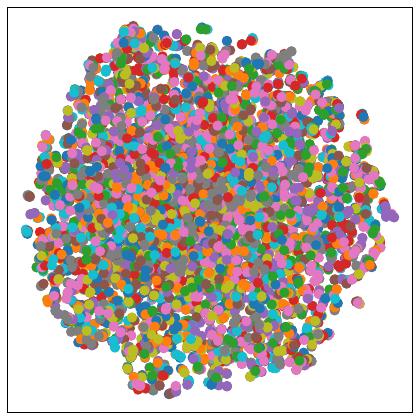}
        \caption{PromptAlign \cite{hassan2024align}}
    \end{subfigure}
    \begin{subfigure}[b]{0.15\textwidth}
        \includegraphics[width=\textwidth]{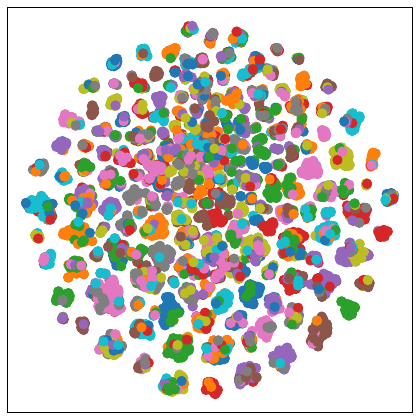}
        \caption{\textbf{TTL} (Ours)}
    \end{subfigure}
    \caption{\textbf{t-SNE visualizations} of the final class embedding from the test sets of $\mathcal{C}_1$ dataset: ImageNet-A, following Table \ref{tab:natural_shift}. TTL could produce linearly separable features for zero-shot generalization than $\mathcal{M}_1$ baselines TPT, and PromptAlign.}
    \label{fig:tsne_small}
\vspace{-0.35cm}
\end{figure}

\begin{figure}[t!]
    \centering
    \begin{subfigure}[b]{0.233\textwidth}
        \includegraphics[width=\textwidth]{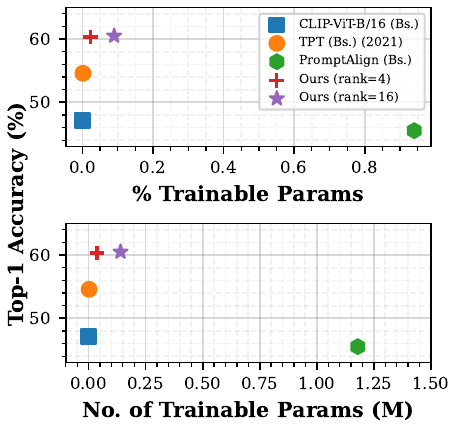}
        \caption{\#Params vs Accuracy}
        \label{fig:train_vs_acc}
    \end{subfigure}
    \begin{subfigure}[b]{0.227\textwidth}
        \includegraphics[width=\textwidth]{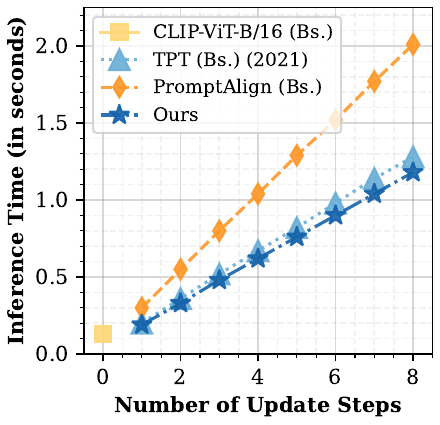}
        \caption{Test-Time Update Steps}
        \label{fig:steps_vs_time}
    \end{subfigure}
        \caption{\textbf{Computational Efficiency.} (a) Trainable Parameters \textit{vs.} Accuracy. Total number of parameters in base CLIP-ViT-B/16 is 124.32 M (b) Number of optimization steps per sample \textit{vs.} the Inference time (in seconds).}
        \label{fig:computation}
\vspace{-0.6cm}
\end{figure}

\vspace{0.10cm}
\noindent\textbf{Trade-off between Inference Efficiency and Performance}: 
We analyze three factors influencing TTL's efficiency and performance: the cutoff percentile $\rho$ for confidence selection, the number of augmented views $N$ at test-time, and the number of update steps $S$. As shown in Figure \ref{fig:cutoff_percent}, TTL achieves maximum performance when $\rho=1$ \ie incorporating the entropy contribution from all the augmented views through our weighted entropy approach. Figure \ref{fig:aug_views} shows a performance gain with increasing $N$, plateauing around $N=64$. Figure \ref{fig:update_steps} shows that with additional update steps, TTL consistently adapts better to the test sample, in contrast to TPT, whose performance trajectory is lower than TTL and exhibits optimal performance with \textit{S}=1, and further updates do not enhance the performance. This suggests that a few optimization steps suffice for optimal generalization.

\vspace{0.1cm}
\noindent\textbf{Trade-off Between Computational Cost and Performance}: TTL introduces trainable parameters (\texttt{TP}) for the attention matrices of the Image Encoder. In comparison, prompt tuning involves 2K \texttt{TP}, and multi-modal tuning employs 1.18M \texttt{TP}, while TTL utilizes only 36K \texttt{TP} for updates during inference. This signifies a subtle trade-off between the number of \texttt{TP} and efficiency, as depicted in Figure \ref{fig:train_vs_acc}. Notably, this additional \texttt{TP} in TTL, compared to TPT and CLIP, does not adversely impact or introduce extra latency during test-time optimization. In fact, as illustrated in Figure \ref{fig:steps_vs_time}, TTL achieves slightly lower inference time with an increasing number of update steps, showcasing superior performance with higher parameter efficiency.


\section{Conclusion}
We present Test-Time Low-rank adaptation (TTL), a novel parameter-efficient strategy for achieving zero-shot generalization in vision-language models (VLMs). TTL provides an efficient alternative to traditional test-time prompt tuning methods by updating the attention weights of CLIP's visual encoder using Low-Rank adapters, thereby adapting the model for downstream recognition tasks without any fine-tuning or pre-training. Additionally, TTL incorporates a confidence maximization mechanism through the utilization of weighted entropy loss derived from augmented sample predictions. Notably, TTL achieves superior performance without the requirement of a source dataset or pre-trained prompts, outperforming current state-of-the-art CLIP zero-shot generalization methods in both domain generalization and cross-dataset evaluation scenarios. We envision that our work will inspire further exploration into optimal strategies for harnessing test-time adaptation in foundational models.

%
%

\bibliographystyle{splncs04}
\bibliography{egbib}

\newpage
\appendix

\noindent\section{\begin{huge} \textbf{Appendix} \vspace{4mm} \end{huge}}

\subsection{Experimental Setup}
\label{appendix:implementation}

\noindent\textbf{Hardware and Software details}: We implement the CLIP ViT-B/16 backbone architecture for implementing our TTL model. TTL and comparative baseline models are executed on a single NVIDIA A100 40GB GPU, leveraging the PyTorch framework.

\vspace{0.10cm}
\noindent\textbf{Reproducibility}:
We conducted additional experiments using the settings outlined in Sec. \ref{sec:exp}. To compare baselines across additional model backbones, we specifically implemented CLIP and TPT. Other methods, such as PromptAlign \cite{hassan2024align}, necessitate computation of source data statistics using respective backbones before inference. Since these source data statistics for backbones other than ViT-B/16 are not provided, and recalculating them across more than a million LAION samples would be computationally expensive, we excluded those comparisons.

\begin{figure}[h]
    \centering
    \begin{subfigure}[b]{0.22\textwidth}
        \includegraphics[width=\textwidth]{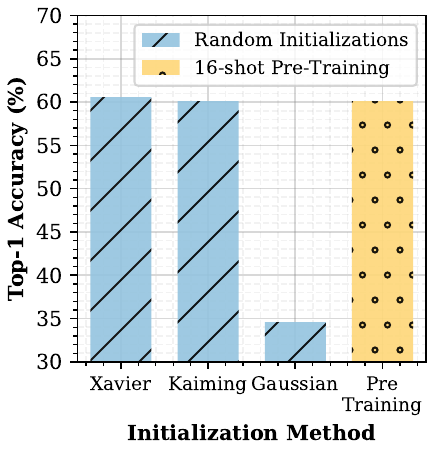}
        \caption{Weight Initializations}
        \vspace{0.2cm}
        \label{fig:init_methods}
    \end{subfigure}
    \begin{subfigure}[b]{0.23\textwidth}
        \includegraphics[width=\textwidth]{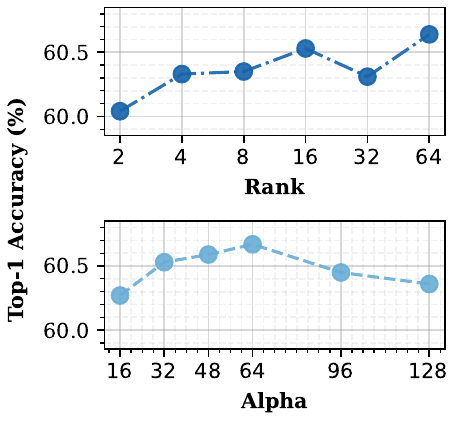}
        \caption{Scaling factors Rank and Alpha of TTL}
        \label{fig:rank_and_alpha}
    \end{subfigure}
    \begin{subfigure}[b]{0.23\textwidth}
        \includegraphics[width=\textwidth]{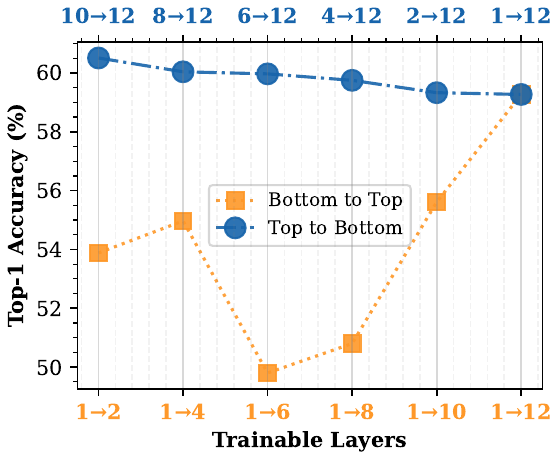}
        \caption{ViT layers to adapt during Test-time model optimization}
        \label{fig:trainable_params}
    \end{subfigure} 
    \caption{Ablating the effects of different TTL components.}
    \label{fig:init_layers_rank}
\vspace{-0.35cm}
\end{figure}

\subsection{Effect of TTL Components}
\label{appendix:ttl_components}


\noindent\textbf{Weight Initialization}: We assess different methods for \textit{initializing LoRA weights} in TTL, including Xavier \cite{glorot2010understanding}, Kaiming \cite{he2015delving}, and Gaussian initializations as shown in Figure \ref{fig:init_methods}. Additionally, we conduct 16-shot pre-training of TTL's adapters using ImageNet dataset following the settings of PromptAlign \cite{hassan2024align}. Our analysis shows that Xavier, Kaiming, and Pre-trained initialized weights exhibit nearly similar performance, with Xavier showing best performance gain of +0.4 compared to Kaiming. Gaussian initialization results in worse performance. Therefore we conclude that weight initialization plays a vital role in the performance gain.

\vspace{0.10cm}
\noindent\textbf{Adaptation Layers}: In Figure \ref{fig:trainable_params}, we investigate the \textit{influence of adapting LoRA adapters in specific layers} of the image encoder on the performance of TTL. Our results reveal that TTL has the most significant impact when adaptation is focused on optimizing attention blocks in the last layers, specifically layers $10 \rightarrow 12$. This is evident as the later layers of transformer based models such as CLIP are more discriminative, capturing high-level representations.

\vspace{0.10cm}
\noindent\textbf{Rank and Alpha}: We investigate the \textit{impact of change in rank} ($r$) \textit{and alpha} ($\alpha$) (Eq. \ref{eqn:lora_standard} of main paper) on the performance. Our observations reveal that with increasing $r$, there is a notable enhancement in performance, as depicted in Figure \ref{fig:rank_and_alpha}. This phenomenon can be attributed to the availability of more trainable parameters within attention groups at higher ranks. Consequently, this suggests that achieving higher performance gains may necessitate sacrificing parameter efficiency.


\subsection{Effect of Entropy Margin} 
In Eq. \ref{eq:tpt_2}, entropy margin $\varepsilon$ is simply a normalization factor to control the sensitivity of the exponentiation to small changes in entropy.  As shown in Figure \ref{fig:wt_entropy_margin}, across different $\varepsilon$ values, a negligible change of 0.04 in performance is observed, indicating that TTL's performance is insensitive towards change in $\varepsilon$ which signfies that TTL can be used with broad range of $\varepsilon$ values.

\begin{figure}[h]
\centering
\includegraphics[width=0.25\textwidth]{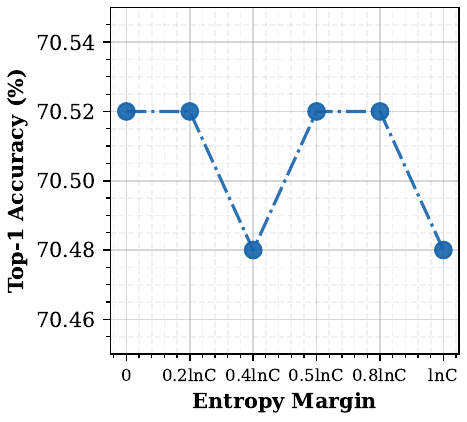}
\caption{Effect of weighted entropy margin. C=$e^x$.}
\label{fig:wt_entropy_margin}
\vspace{-0.35cm}
\end{figure}



{\renewcommand{\arraystretch}{1.0}
\begin{table*}[t]
\caption{Analysis on the components of TTL and different methods for test-time optimization of Vision-Language Model. ${\dag}$ denotes 16-shot pre-training and * denotes reported accuracy. \textbf{Top-1 Accuracy} indicates ImageNet-A accuracy. The arrow ${\color{myforestgreen}\uparrow}$ and ${\color{red}\downarrow}$ indicate \textbf{improvements} and \textbf{decrements} of our method against the  CLIP ($bs.$) method, \ie, CLIP-ViT-B/16. }

\centering
\resizebox{\textwidth}{!}{%
\begin{tabular}{l|ccccc|c} \hline
\textbf{Method} & \textbf{~~Tunable Parameter~~} & \textbf{Entropy-Loss} & \textbf{NO Pre-training} & \textbf{NO Auxiliary-Data} & \textbf{NO External-Model} & \textbf{Top-1 Accuracy} \\ 
\cmidrule(r){1-1} \cmidrule(lr){2-2} \cmidrule(r){3-3} \cmidrule(r){4-4} \cmidrule(r){5-5} \cmidrule(r){6-6} \cmidrule(l){7-7}
\rowcolor{mygray} 
CLIP\stdvuno{$bs.$} & {--} & {--} & {--} & {--} & {--} & 47.14\stdvuno{$bs.$} \\

\rowcolor[HTML]{F6FBFF} 
TPT \cite{shu2022test} & {Text Prompt} & {\color[HTML]{4D5156} \textbf{\Checkmark}} & {\Checkmark} & {\Checkmark} & {\Checkmark} & 54.59\stdvun{7.45} \\

\rowcolor[HTML]{eff6fc} 
DiffTPT* \cite{feng2023diverse} & {Text Prompt} & {\color[HTML]{4D5156} \textbf{\Checkmark}} & {\Checkmark} & {\Checkmark} & {\color[HTML]{4D5156} \textbf{\ding{56}}} & 55.68\stdvun{8.54} \\

\rowcolor[HTML]{e7f3fa} 
PromptAlign \cite{hassan2024align} & {Multi-Modal Prompt} & {\color[HTML]{4D5156} \textbf{\Checkmark}} & {\Checkmark} & {\color[HTML]{4D5156} \textbf{\ding{56}}} & {\Checkmark} & 45.52\stdvuru{1.62} \\

\rowcolor[HTML]{e3f2ff} 
PromptAlign$^{\dag}$ \cite{hassan2024align} & {Multi-Modal Prompt} & {\color[HTML]{4D5156} \textbf{\Checkmark}} & {\color[HTML]{4D5156} \textbf{\ding{56}}} & {\color[HTML]{4D5156} \textbf{\ding{56}}} & {\Checkmark} & 59.03\stdvun{11.89} \\

\rowcolor[HTML]{deefff} 
\textbf{TTL (Ours)} & {Low-Rank Attentions} & {\color[HTML]{4D5156} \textbf{\Checkmark}} & {\Checkmark} & {\Checkmark} & {\Checkmark} & \textbf{60.51}\stdvun{13.37} \\ \hline

\end{tabular}}
\label{tab:tick_methods}
\end{table*}

{\renewcommand{\arraystretch}{1.1}
\begin{table*}[t]
\caption{\textbf{Top 1 accuracy} $\%$ of state-of-the-art baselines under \texttt{strict zero-shot settings}, where \textbf{ImageNet-Sk.} indicates the ImageNet-Sketch dataset, \textbf{OOD Avg.} indicates the OOD average results. {\cellcolor{mygray}{$bs.$}} indicates the baseline, \ie, CLIP-ViT-B/32. 
}
\centering
\resizebox{\textwidth}{!}{%
\begin{tabular}{l|ccccc|c|c}
\hline
\textbf{Method} & \textbf{ImageNet} & \textbf{ImageNet-A} & \textbf{ImageNet-V2} & \textbf{ImageNet-R} & \textbf{ImageNet-Sk.} & \textbf{Average} & \textbf{OOD Avg.} \\ 
\cmidrule(r){1-1} \cmidrule(lr){2-2} \cmidrule(lr){3-3} \cmidrule(lr){4-4} \cmidrule(lr){5-5} \cmidrule(lr){6-6} \cmidrule(lr){7-7} \cmidrule(l){8-8}

\rowcolor{mygray} 
CLIP-ViT-B/32 & 59.63\stdvuno{$bs.$} & 29.57\stdvuno{$bs.$} & 54.74\stdvuno{$bs.$} & 66.27\stdvuno{$bs.$} & 40.77\stdvuno{$bs.$} & 50.20\stdvuno{$bs.$} & 47.84\stdvuno{$bs.$} \\

\rowcolor[HTML]{e7f3fa} 
TPT${\color{cyan}_{2022}}$\cite{shu2022test} & 60.93\stdvun{1.30} & 34.61\stdvun{5.04} & 57.15\stdvun{2.41} & 69.60\stdvun{3.33} & 41.62\stdvun{0.85} & 52.78\stdvun{2.58} & 50.75\stdvun{2.91} \\

\rowcolor[HTML]{deefff} 
\textbf{TTL (Ours)} & \textbf{66.94}\stdvun{7.31} & \textbf{42.43}\stdvun{12.86} & \textbf{58.85}\stdvun{4.11} & \textbf{71.29}\stdvun{5.02} & \textbf{43.54}\stdvun{2.77} & \textbf{56.61}\stdvun{6.41} & \textbf{54.03}\stdvun{6.19} \\ \hline

\end{tabular}
}
\label{tab:different_backbone}
\end{table*}

\subsection{Different Backbone Scales}
\label{appendix:backbone}
To verify the scalability of approach, We conduct additional experiments by replacing the ViT-B/16 backbone with ViT-B/32 variant of CLIP. We use same hyperparameters as in original TTL settings without undergoing any hyperparameter tuning. As shown in Table \ref{tab:different_backbone}, TTL consistently outperforms CLIP, TPT, and other baselines even with different backbone.
In terms of in-domain generalization accuracy, with an accuracy of \textbf{56.61}, TTL achieves an average gain of +6.41 and +3.83 compared to CLIP and TPT, respectively. Across out-of-distribution (OOD) shift datasets, TTL demonstrates an average accuracy of \textbf{54.03}, with specific gains of  +6.19 and +3.28 over CLIP and TPT respectively.






\subsection{Qualitative Analysis on Feature Shift}
\label{appendix:tsne}

The t-SNE visualizations of various approaches, extracted from visual token embeddings of the last layer of the visual encoder on the ImageNet-A dataset, are presented in Figure \ref{fig:tsne}. It is evident that TPT \cite{shu2022test} exhibits a similar arrangement in the t-SNE visualization to that of CLIP \cite{radford2021learning}, as TPT updates prompts solely on the textual side without modifying visual token embeddings. Conversely, PromptAlign \cite{hassan2024align}, which aligns the visual embeddings of test samples to a given source statistic, fails to achieve a optimal class separation boundary, even with pre-trained prompts.
In contrast, TTL demonstrates clearly separable boundaries across various hyperparameters like rank $r$, indicating effective model adaptation for a given test sample.

\begin{figure*}[t]
    \centering
    \begin{subfigure}[b]{0.28\textwidth}
        \includegraphics[width=\textwidth]{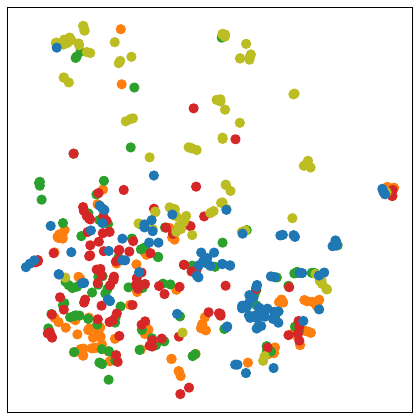}
        \caption{CLIP \cite{radford2021learning}}
    \end{subfigure}
    \begin{subfigure}[b]{0.28\textwidth}
        \includegraphics[width=\textwidth]{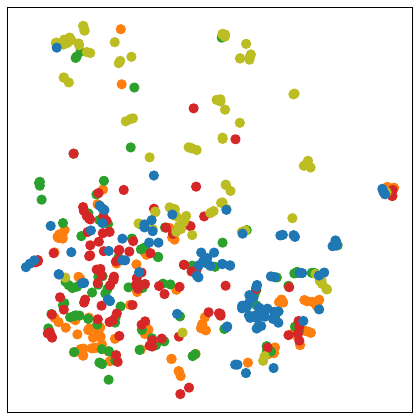}
        \caption{TPT \cite{shu2022test}}
    \end{subfigure}
    \begin{subfigure}[b]{0.28\textwidth}
        \includegraphics[width=\textwidth]{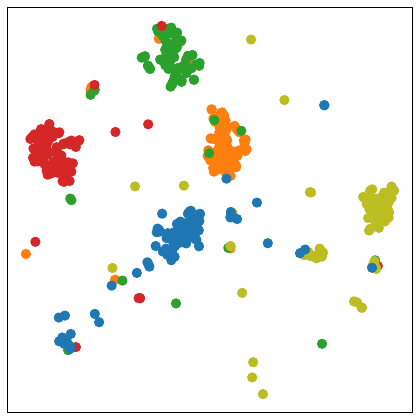}
        \caption{\textbf{TTL} (rank $r=4$)}
    \end{subfigure}
    
    \begin{subfigure}[b]{0.28\textwidth}
        \includegraphics[width=\textwidth]{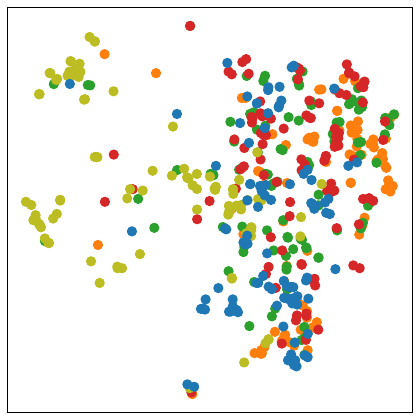}
        \caption{PromptAlign \cite{hassan2024align}}
    \end{subfigure}
    \begin{subfigure}[b]{0.28\textwidth}
        \includegraphics[width=\textwidth]{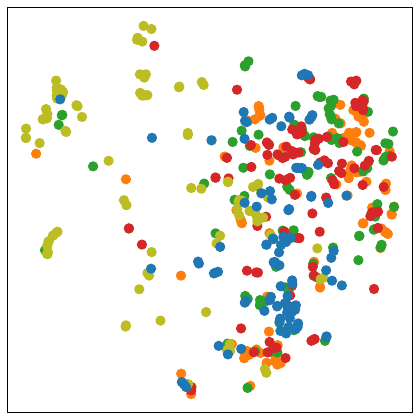}
        \caption{PromptAlign$^\dagger$ \cite{hassan2024align}}
    \end{subfigure}
    \begin{subfigure}[b]{0.28\textwidth}
        \includegraphics[width=\textwidth]{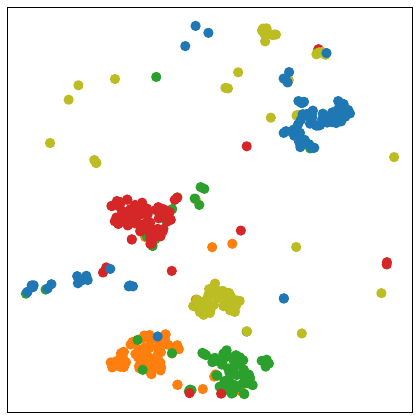}
        \caption{\textbf{TTL} (rank $r=16$)}
    \end{subfigure}
    
    \begin{subfigure}[b]{0.28\textwidth}
        \includegraphics[width=\textwidth]{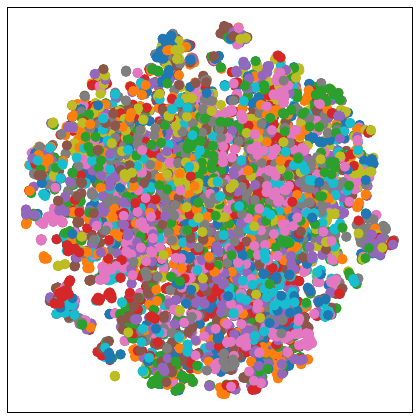}
        \caption{CLIP \cite{radford2021learning}}
    \end{subfigure}
    \begin{subfigure}[b]{0.28\textwidth}
        \includegraphics[width=\textwidth]{Figures/tpt_features_A_allclasses.pdf}
        \caption{TPT \cite{shu2022test}}
    \end{subfigure}
    \begin{subfigure}[b]{0.28\textwidth}
        \includegraphics[width=\textwidth]{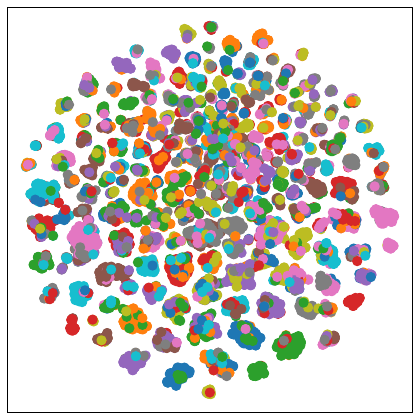}
        \caption{\textbf{TTL} (rank $r=4$)}
    \end{subfigure}
    
    \begin{subfigure}[b]{0.28\textwidth}
        \includegraphics[width=\textwidth]{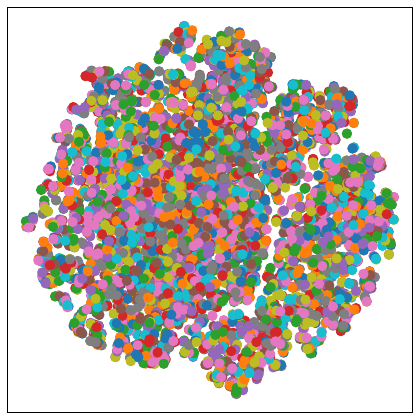}
        \caption{PromptAlign \cite{hassan2024align}}
    \end{subfigure}
    \begin{subfigure}[b]{0.28\textwidth}
        \includegraphics[width=\textwidth]{Figures/promptalign_P_features_A_all_classes.pdf}
        \caption{PromptAlign$^\dagger$ \cite{hassan2024align}}
    \end{subfigure}
    \begin{subfigure}[b]{0.28\textwidth}
        \includegraphics[width=\textwidth]{Figures/ttl_features_A_allclasses.pdf}
        \caption{\textbf{TTL} (rank $r=16$)}
    \end{subfigure}
    
    \caption{\textbf{t-SNE visualizations} of the final class embedding from the test sets of $\mathcal{C}_1$ dataset: ImageNet-A, following Table \ref{tab:natural_shift}. TTL could produce linearly separable features for zero-shot generalization than $\mathcal{M}_1$ baselines like CLIP, TPT, and PromptAlign. $\dagger$ indicates 16-shot pre-training. \textbf{(a)} to \textbf{(f)} represents 5 classes of ImageNet-A with highest number of samples, while \textbf{(g)} to \textbf{(l)} represents all 200 classes of ImageNet-A.}
    \label{fig:tsne}
\end{figure*}

\subsection{Qualitative Analysis on Attention Shift}
\label{appendix:attn}
TTL demonstrates enhanced attention towards target discriminative visual areas while simultaneously reducing attention towards background regions when compared to CLIP. This underscores the direct influence of TTL's LoRA parameters in updating the query and value weights of VLM's attention blocks, resulting in improved predictions by focusing the model's attention on object of interest. As shown in Figure \ref{fig:attention_maps}, TTL amplifies attention towards the object of interest while attenuating attention towards background areas across input samples. By adapting to task-specific data, TTL effectively identifies relevant features in the input sample, thereby improving prediction accuracy.

\begin{figure*}[t]
    \centering

    \begin{subfigure}[b]{0.135\textwidth}
        \caption*{Input Image}
    \end{subfigure}
    \begin{subfigure}[b]{0.135\textwidth}
        \caption*{CLIP \cite{radford2021learning}}
    \end{subfigure}
    \begin{subfigure}[b]{0.135\textwidth}
        \caption*{\textbf{TTL} (Ours)}
    \end{subfigure}
    \hspace{0.03\textwidth}
    \begin{subfigure}[b]{0.135\textwidth}
        \caption*{Input Image}
    \end{subfigure}
    \begin{subfigure}[b]{0.135\textwidth}
        \caption*{CLIP \cite{radford2021learning}}
    \end{subfigure}
    \begin{subfigure}[b]{0.135\textwidth}
        \caption*{\textbf{TTL} (Ours)}
    \end{subfigure}
    
    \begin{subfigure}[b]{0.135\textwidth}
        \frame{\includegraphics[width=\textwidth]{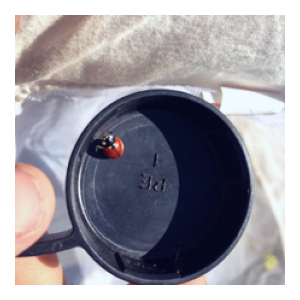}}
    \end{subfigure}
    \begin{subfigure}[b]{0.135\textwidth}
        \frame{\includegraphics[width=\textwidth]{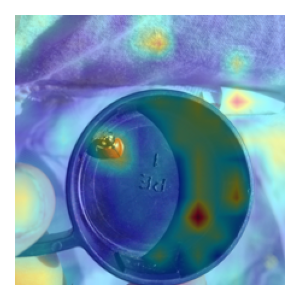}}
    \end{subfigure}
    \begin{subfigure}[b]{0.135\textwidth}
        \frame{\includegraphics[width=\textwidth]{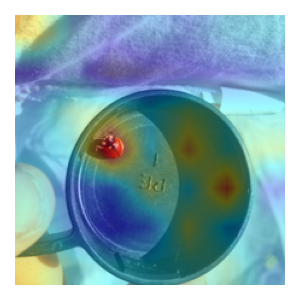}}
    \end{subfigure}
    \hspace{0.03\textwidth}
    \begin{subfigure}[b]{0.135\textwidth}
        \frame{\includegraphics[width=\textwidth]{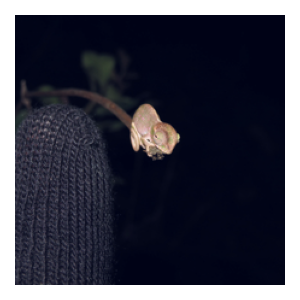}}
    \end{subfigure}
    \begin{subfigure}[b]{0.135\textwidth}
        \frame{\includegraphics[width=\textwidth]{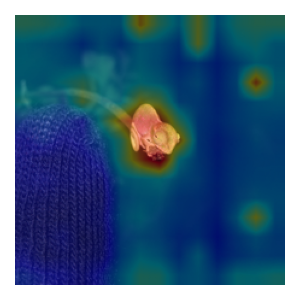}}
    \end{subfigure}
    \begin{subfigure}[b]{0.135\textwidth}
        \frame{\includegraphics[width=\textwidth]{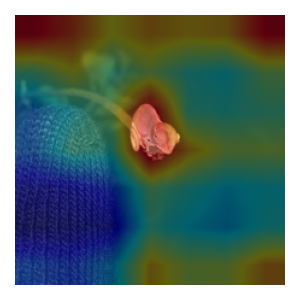}}
    \end{subfigure}
    
    \begin{subfigure}[b]{0.135\textwidth}
        \frame{\includegraphics[width=\textwidth]{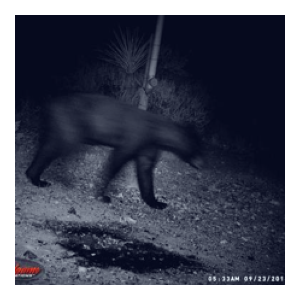}}
    \end{subfigure}
    \begin{subfigure}[b]{0.135\textwidth}
        \frame{\includegraphics[width=\textwidth]{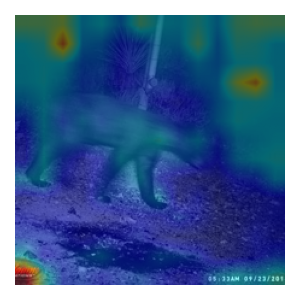}}
    \end{subfigure}
    \begin{subfigure}[b]{0.135\textwidth}
        \frame{\includegraphics[width=\textwidth]{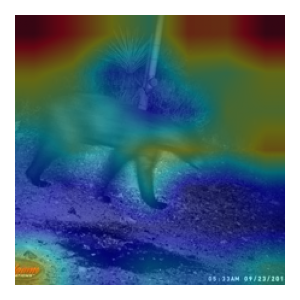}}
    \end{subfigure}
    \hspace{0.03\textwidth}
    \begin{subfigure}[b]{0.135\textwidth}
        \frame{\includegraphics[width=\textwidth]{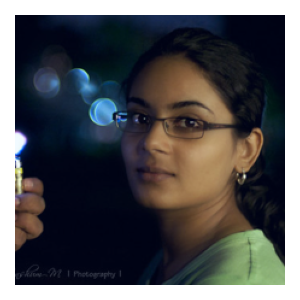}}
    \end{subfigure}
    \begin{subfigure}[b]{0.135\textwidth}
        \frame{\includegraphics[width=\textwidth]{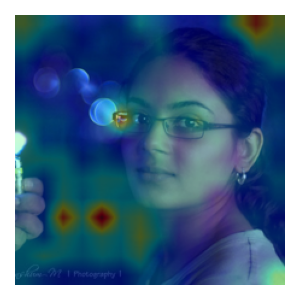}}
    \end{subfigure}
    \begin{subfigure}[b]{0.135\textwidth}
        \frame{\includegraphics[width=\textwidth]{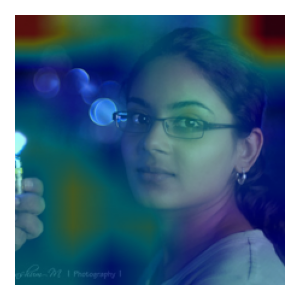}}
    \end{subfigure}

    \begin{subfigure}[b]{0.135\textwidth}
        \frame{\includegraphics[width=\textwidth]{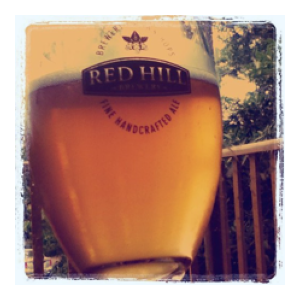}}
    \end{subfigure}
    \begin{subfigure}[b]{0.135\textwidth}
        \frame{\includegraphics[width=\textwidth]{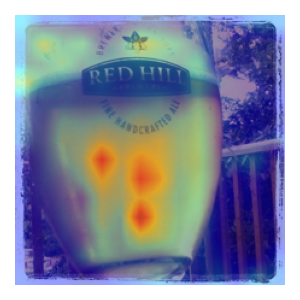}}
    \end{subfigure}
    \begin{subfigure}[b]{0.135\textwidth}
        \frame{\includegraphics[width=\textwidth]{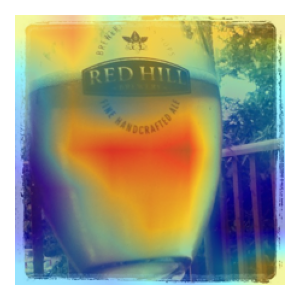}}
    \end{subfigure}
    \hspace{0.03\textwidth}
    \begin{subfigure}[b]{0.135\textwidth}
        \frame{\includegraphics[width=\textwidth]{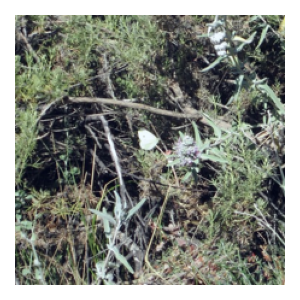}}
    \end{subfigure}
    \begin{subfigure}[b]{0.135\textwidth}
        \frame{\includegraphics[width=\textwidth]{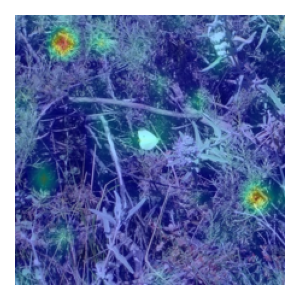}}
    \end{subfigure}
    \begin{subfigure}[b]{0.135\textwidth}
        \frame{\includegraphics[width=\textwidth]{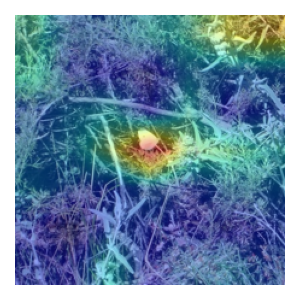}}
    \end{subfigure}

    \begin{subfigure}[b]{0.135\textwidth}
        \frame{\includegraphics[width=\textwidth]{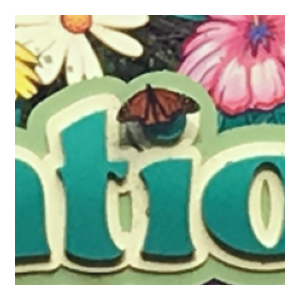}}
    \end{subfigure}
    \begin{subfigure}[b]{0.135\textwidth}
        \frame{\includegraphics[width=\textwidth]{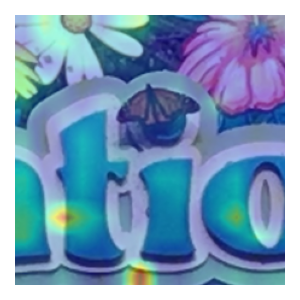}}
    \end{subfigure}
    \begin{subfigure}[b]{0.135\textwidth}
        \frame{\includegraphics[width=\textwidth]{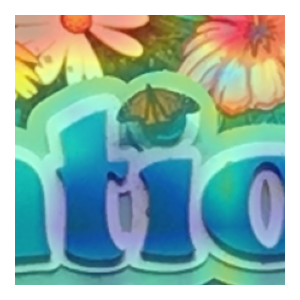}}
    \end{subfigure}
    \hspace{0.03\textwidth}
    \begin{subfigure}[b]{0.135\textwidth}
        \frame{\includegraphics[width=\textwidth]{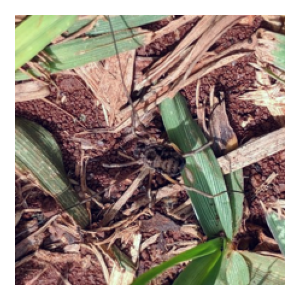}}
    \end{subfigure}
    \begin{subfigure}[b]{0.135\textwidth}
        \frame{\includegraphics[width=\textwidth]{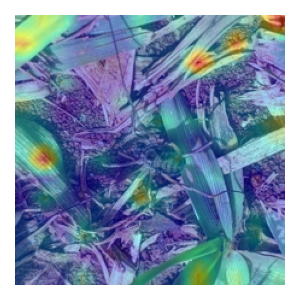}}
    \end{subfigure}
    \begin{subfigure}[b]{0.135\textwidth}
        \frame{\includegraphics[width=\textwidth]{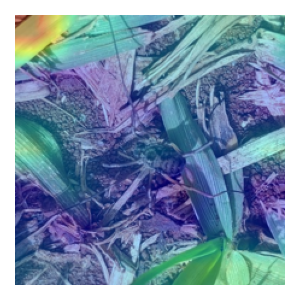}}
    \end{subfigure}

    \begin{subfigure}[b]{0.135\textwidth}
        \frame{\includegraphics[width=\textwidth]{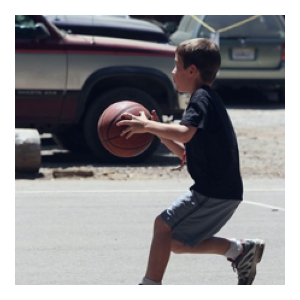}}
    \end{subfigure}
    \begin{subfigure}[b]{0.135\textwidth}
        \frame{\includegraphics[width=\textwidth]{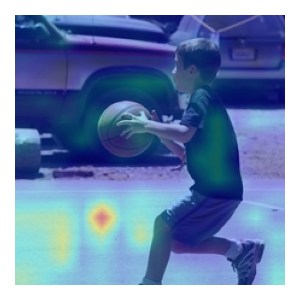}}
    \end{subfigure}
    \begin{subfigure}[b]{0.135\textwidth}
        \frame{\includegraphics[width=\textwidth]{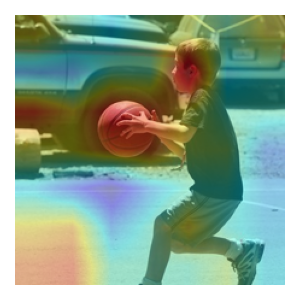}}
    \end{subfigure}
    \hspace{0.03\textwidth}
    \begin{subfigure}[b]{0.135\textwidth}
        \frame{\includegraphics[width=\textwidth]{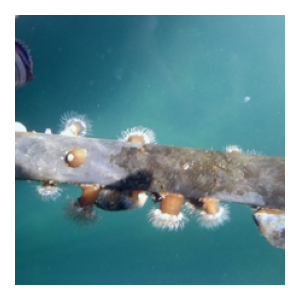}}
    \end{subfigure}
    \begin{subfigure}[b]{0.135\textwidth}
        \frame{\includegraphics[width=\textwidth]{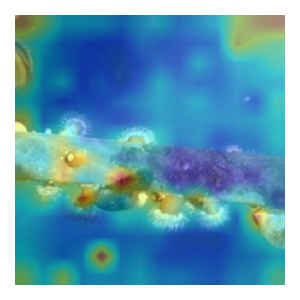}}
    \end{subfigure}
    \begin{subfigure}[b]{0.135\textwidth}
        \frame{\includegraphics[width=\textwidth]{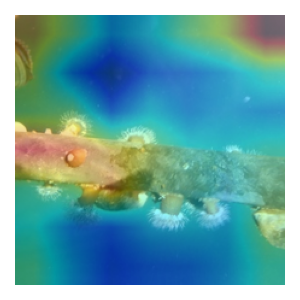}}
    \end{subfigure}

    \begin{subfigure}[b]{0.135\textwidth}
        \frame{\includegraphics[width=\textwidth]{Figures/grad_cams/imgs/img_27.png}}
    \end{subfigure}
    \begin{subfigure}[b]{0.135\textwidth}
        \frame{\includegraphics[width=\textwidth]{Figures/grad_cams/clip_bs/clip_bs_attn_27.png}}
    \end{subfigure}
    \begin{subfigure}[b]{0.135\textwidth}
        \frame{\includegraphics[width=\textwidth]{Figures/grad_cams/ttl/ttl_attn_27.png}}
    \end{subfigure}
    \hspace{0.03\textwidth}
    \begin{subfigure}[b]{0.135\textwidth}
        \frame{\includegraphics[width=\textwidth]{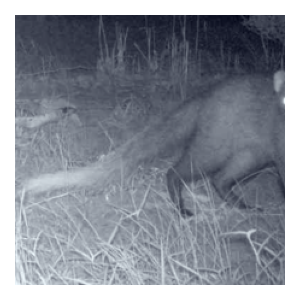}}
    \end{subfigure}
    \begin{subfigure}[b]{0.135\textwidth}
        \frame{\includegraphics[width=\textwidth]{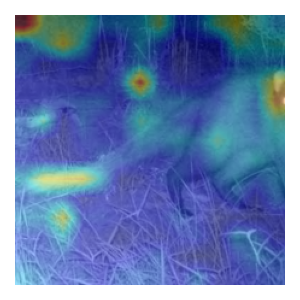}}
    \end{subfigure}
    \begin{subfigure}[b]{0.135\textwidth}
        \frame{\includegraphics[width=\textwidth]{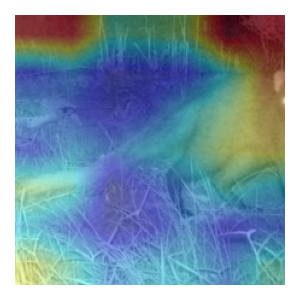}}
    \end{subfigure}

    \begin{subfigure}[b]{0.135\textwidth}
        \frame{\includegraphics[width=\textwidth]{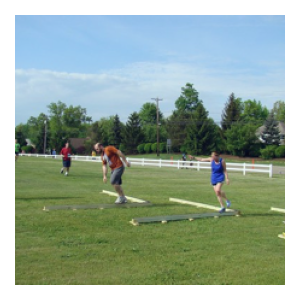}}
    \end{subfigure}
    \begin{subfigure}[b]{0.135\textwidth}
        \frame{\includegraphics[width=\textwidth]{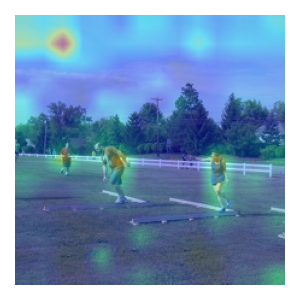}}
    \end{subfigure}
    \begin{subfigure}[b]{0.135\textwidth}
        \frame{\includegraphics[width=\textwidth]{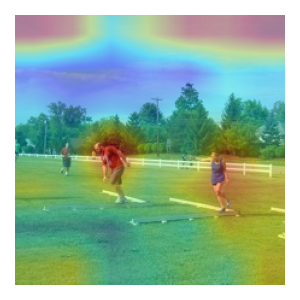}}
    \end{subfigure}
    \hspace{0.03\textwidth}
    \begin{subfigure}[b]{0.135\textwidth}
        \frame{\includegraphics[width=\textwidth]{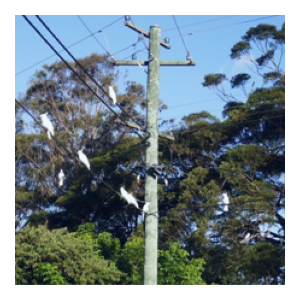}}
    \end{subfigure}
    \begin{subfigure}[b]{0.135\textwidth}
        \frame{\includegraphics[width=\textwidth]{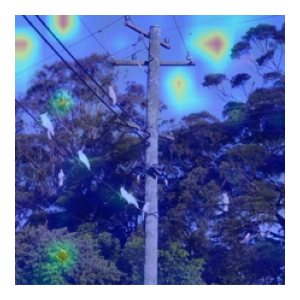}}
    \end{subfigure}
    \begin{subfigure}[b]{0.135\textwidth}
        \frame{\includegraphics[width=\textwidth]{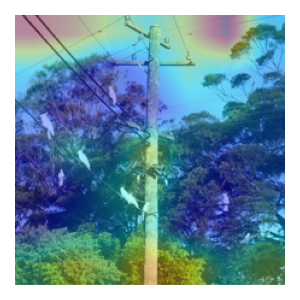}}
    \end{subfigure}

    \begin{subfigure}[b]{0.135\textwidth}
        \frame{\includegraphics[width=\textwidth]{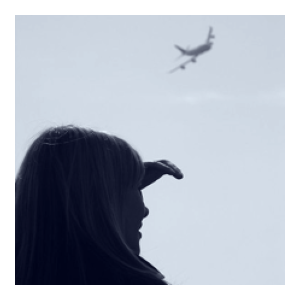}}
    \end{subfigure}
    \begin{subfigure}[b]{0.135\textwidth}
        \frame{\includegraphics[width=\textwidth]{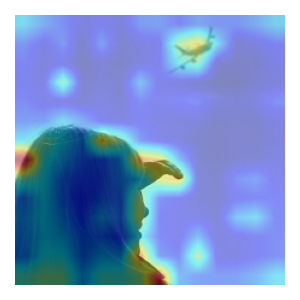}}
    \end{subfigure}
    \begin{subfigure}[b]{0.135\textwidth}
        \frame{\includegraphics[width=\textwidth]{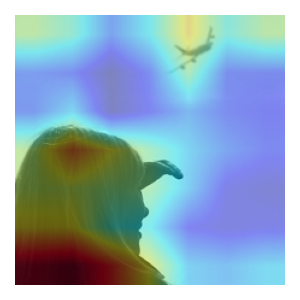}}
    \end{subfigure}
    \hspace{0.03\textwidth}
    \begin{subfigure}[b]{0.135\textwidth}
        \frame{\includegraphics[width=\textwidth]{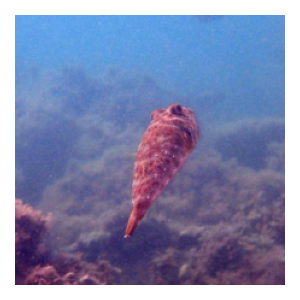}}
    \end{subfigure}
    \begin{subfigure}[b]{0.135\textwidth}
        \frame{\includegraphics[width=\textwidth]{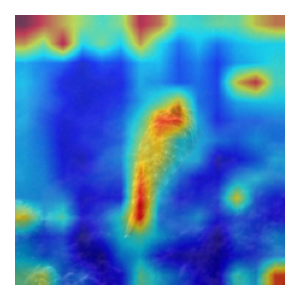}}
    \end{subfigure}
    \begin{subfigure}[b]{0.135\textwidth}
        \frame{\includegraphics[width=\textwidth]{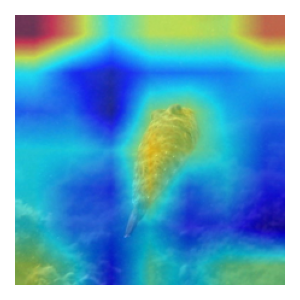}}
    \end{subfigure}

    \begin{subfigure}[b]{0.135\textwidth}
        \frame{\includegraphics[width=\textwidth]{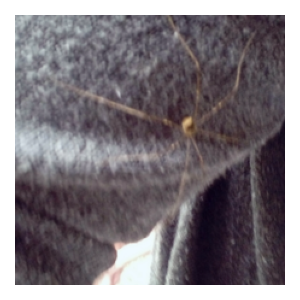}}
    \end{subfigure}
    \begin{subfigure}[b]{0.135\textwidth}
        \frame{\includegraphics[width=\textwidth]{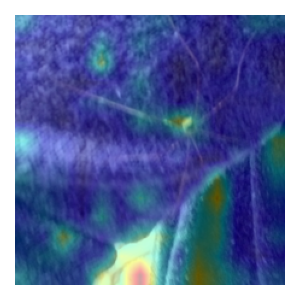}}
    \end{subfigure}
    \begin{subfigure}[b]{0.135\textwidth}
        \frame{\includegraphics[width=\textwidth]{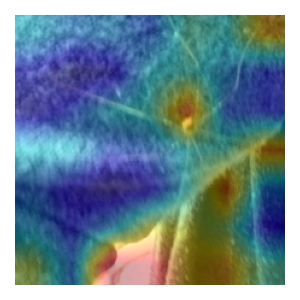}}
    \end{subfigure}
    \hspace{0.03\textwidth}
    \begin{subfigure}[b]{0.135\textwidth}
        \frame{\includegraphics[width=\textwidth]{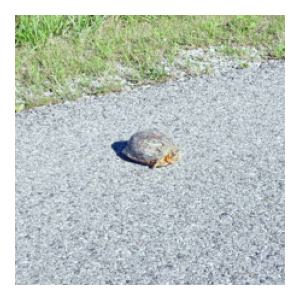}}
    \end{subfigure}
    \begin{subfigure}[b]{0.135\textwidth}
        \frame{\includegraphics[width=\textwidth]{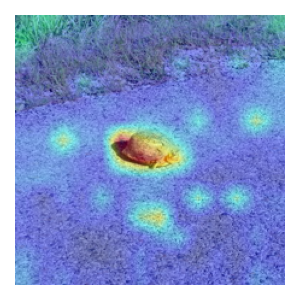}}
    \end{subfigure}
    \begin{subfigure}[b]{0.135\textwidth}
        \frame{\includegraphics[width=\textwidth]{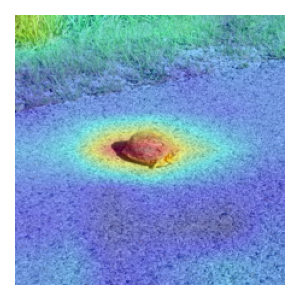}}
    \end{subfigure}

    \caption{\textbf{Attention map visualizations.} TTL updates the attention weights to prioritize features that are more relevant for the target task to better represent domain-specific features, whereas pre-trained CLIP shows inadequacy in capturing such features.}
    \label{fig:attention_maps}
\end{figure*}

\subsection{Limitations and Future Directions}
\noindent\textbf{Limitations}:
While TTL doesn't necessitate any source data or annotations, our approach does involve a one-step backpropagation process when adapting the low-rank weights during testing. As TTL generates multiple augmented views of a single test sample, it leads to higher memory usage during inference compared to the foundational CLIP model.

\vspace{0.10cm}
\noindent\textbf{Future Research Directions}: 
We present several directions for future works. 
\vspace{-0.20cm}
\begin{itemize}[left=0pt]
    \item The concept of TTL has the potential to be extended to various downstream tasks like segmentation and detection, thereby enhancing their ability for zero-shot generalization.
    \vspace{-0.20cm}
    \item Exploring methods to minimize the memory overhead of TTL and enhance its computational efficiency would be interesting.
    \vspace{-0.20cm}
    \item TTL could also be adapted for other domain-specific classification and visual reasoning tasks such as medical imaging and remote sensing applications.
    \vspace{-0.20cm}
    \item A promising direction for further research would involve evaluating and devising strategies to enhance the adversarial robustness of vision-language foundational models built upon TTL.
\end{itemize}

{\renewcommand{\arraystretch}{1.0}
\begin{table*}[b]
\caption{\textbf{Weighted Entropy on other baselines.} \textit{w} = "with". PromptAlign$^{\dag}$ indicates using pre-trained weights.}

\centering
\resizebox{\textwidth}{!}{%
\begin{tabular}{l|cccccc} \hline
Method & ~\textbf{Flower102}\cite{nilsback2008automated}~ & ~\textbf{DTD}\cite{cimpoi2014describing}~ & ~\textbf{OxfordPets}\cite{parkhi2012cats}~ & ~\textbf{UCF}\cite{soomro2012dataset}~ & ~\textbf{Caltech101}\cite{fei2004learning}~ & ~\textbf{Aircraft}\cite{maji2013fine}~ \\
\cmidrule(r){1-1} \cmidrule(lr){2-2} \cmidrule(r){3-3} \cmidrule(r){4-4} \cmidrule(r){5-5} \cmidrule(l){6-6} \cmidrule(l){7-7}
\rowcolor{mygray} 
TPT & 69.31 & 46.23 & 86.49 & 66.44 & 92.49 & \textbf{24.90}\\
\rowcolor{mygray} 
TPT \textit{w} Wt. Ent. & 69.56 & 46.69 & 88.58 & 69.18 & 93.55 & 23.14 \\
\rowcolor[HTML]{eff6fc} 
PromptAlign & 51.60 & 27.60 & 75.82 & 57.31 & 87.18 & 6.96 \\
\rowcolor[HTML]{eff6fc} 
PromptAlign \textit{w} Wt. Ent. & 52.05 & 27.66 & 75.94 & 58.10 & 87.61 & 7.10 \\ 
\rowcolor[HTML]{e7f3fa} 
PromptAlign$^{\dag}$ & 70.56 & 45.57 & 88.96 & 69.10 & 92.86 & 23.70 \\
\rowcolor[HTML]{e7f3fa} 
PromptAlign$^{\dag}$ \textit{w} Wt. Ent. & 71.74 & 45.04 & \textbf{89.07} & 68.70 & 93.47 & 24.15\\ 
\rowcolor[HTML]{deefff} 
\textbf{TTL (Ours)} & \textbf{70.48} & \textbf{46.69} & 88.72 & \textbf{69.20} & \textbf{93.63} & 23.82\\ \hline
\end{tabular}
}
\centering
\resizebox{\textwidth}{!}{%
\begin{tabular}{l|cccc|c} \hline
Method & ~~~\textbf{EuroSAT}\cite{helber2019eurosat}~~~ & ~~~\textbf{StanfordCars}\cite{krause20133d}~~~ & ~~~\textbf{Food101}\cite{bossard2014food}~~~ & ~~~\textbf{SUN397}\cite{xiao2010sun}~~~ & ~~~~~\textbf{Average}~~~~~ \\ 
\cmidrule(r){1-1} \cmidrule(lr){2-2} \cmidrule(r){3-3} \cmidrule(r){4-4} \cmidrule(r){5-5} \cmidrule(l){6-6} 
\rowcolor{mygray} 
TPT & 37.15 & 66.50 & 86.93 & 63.48 & 63.99\\
\rowcolor{mygray} 
TPT \textit{w} Wt. Ent. & 41.96 & 66.37 & 84.92 & 64.96 & 64.89 \\
\rowcolor[HTML]{eff6fc}
PromptAlign & 35.57 & 58.70 & 82.23 & 57.84 & 54.08 \\
\rowcolor[HTML]{eff6fc}
PromptAlign \textit{w} Wt. Ent. & 37.74 & 57.99 & 82.15 & 57.98 & 54.43 \\ 
\rowcolor[HTML]{e7f3fa}
PromptAlign$^{\dag}$ & 34.91 & 67.43 & 86.85 & 67.73 & 64.76 \\
\rowcolor[HTML]{e7f3fa}
PromptAlign$^{\dag}$ \textit{w} Wt. Ent. & 36.56 & 67.35 & \textbf{86.91} & \textbf{68.03} & 65.10 \\
\rowcolor[HTML]{deefff}
\textbf{TTL (Ours)} & \textbf{42.02} & \textbf{67.96} & 85.05 & {66.32} & \textbf{65.39}\\ \hline
\end{tabular}
}
\vspace{-0.8em}

\label{tab:weighted_entropy_rebut}
\end{table*}

\begin{figure*}[b]
    \centering
    \begin{subfigure}[b]{0.2475\textwidth}
        \includegraphics[width=\textwidth]{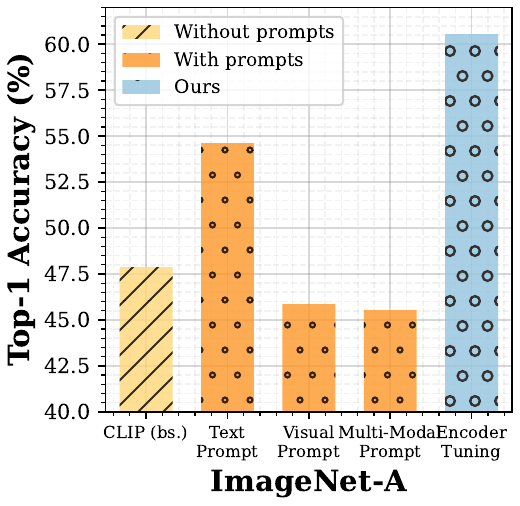}
    \end{subfigure}
    \hfill
    \begin{subfigure}[b]{0.2475\textwidth}
        \includegraphics[width=\textwidth]{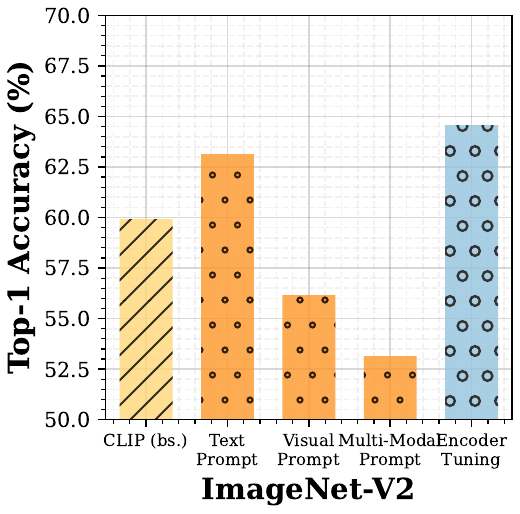}
    \end{subfigure}
    \hfill
    \begin{subfigure}[b]{0.24\textwidth}
        \includegraphics[width=\textwidth]{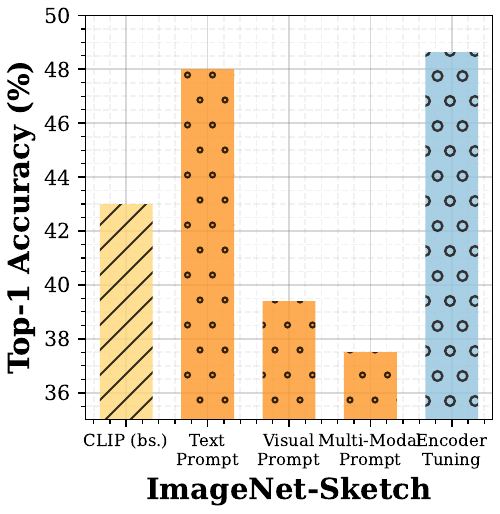}
    \end{subfigure}
    \hfill
    \begin{subfigure}[b]{0.24\textwidth}
        \includegraphics[width=\textwidth]{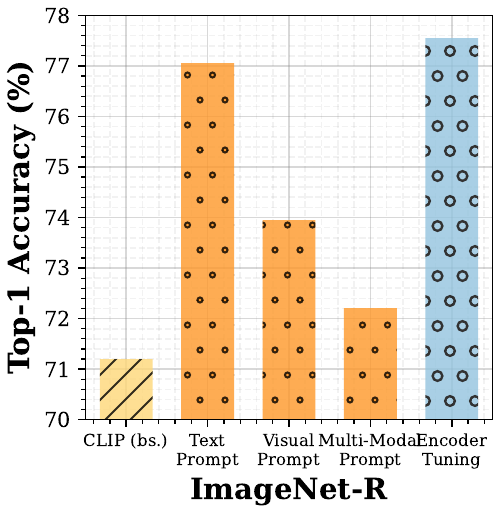}
    \end{subfigure}

    \begin{subfigure}[b]{0.24\textwidth}
        \includegraphics[width=\textwidth]{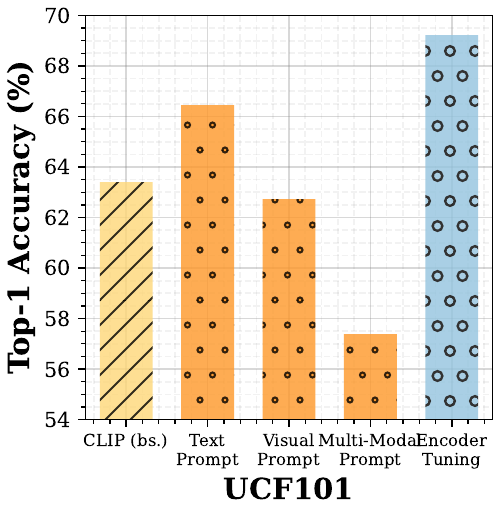}
    \end{subfigure}
    \hfill
    \begin{subfigure}[b]{0.24\textwidth}
        \includegraphics[width=\textwidth]{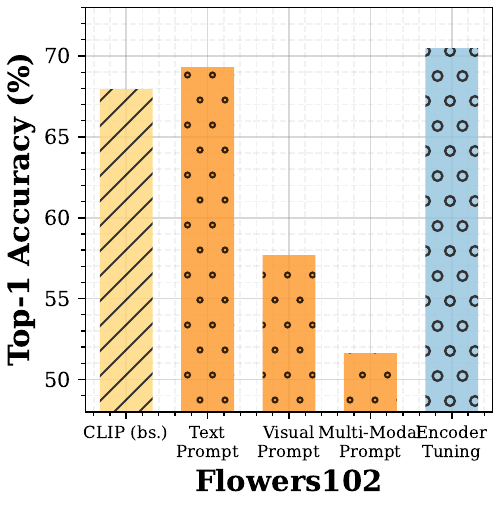}
    \end{subfigure}
    \hfill
    \begin{subfigure}[b]{0.24\textwidth}
        \includegraphics[width=\textwidth]{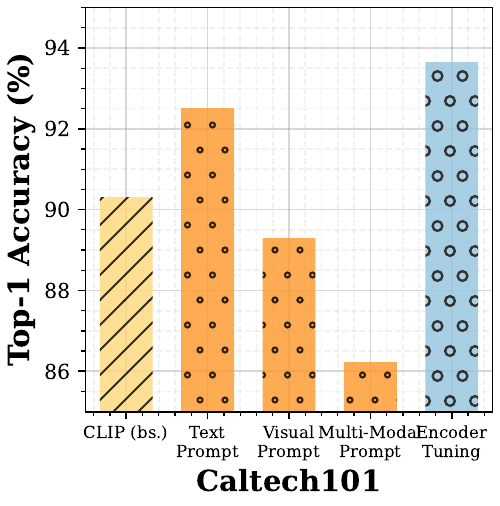}
    \end{subfigure}
    \hfill
    \begin{subfigure}[b]{0.24\textwidth}
        \includegraphics[width=\textwidth]{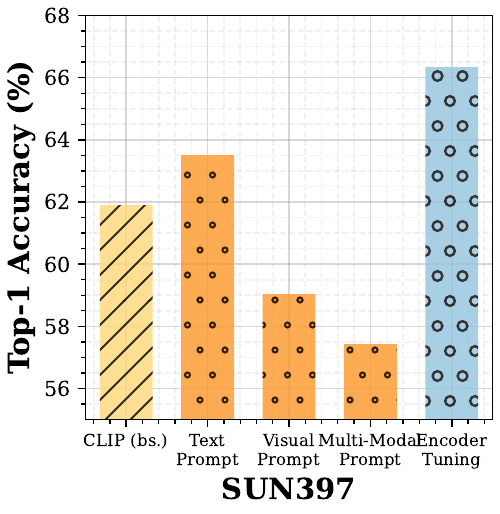}
    \end{subfigure}
    \caption{\textbf{Test-time performance of zero-shot generalization methods.} CLIP \textit{vs.} Textual Prompt Tuning (TPT) \textit{vs.} Visual Prompt Tuning \textit{vs.} Multi-modal Prompt Tuning \textit{vs.} \textbf{TTL (Ours)}. First row denotes the OOD $\mathcal{C}_1$ datasets while Second row denotes 4 Cross-domain $\mathcal{C}_2$ datasets. 
    }
    \label{fig:maple_compare}
\end{figure*}

\end{document}